\documentclass[11pt,a4paper]{article}
\usepackage[hyperref]{emnlp-ijcnlp-2019}
\usepackage{times}
\usepackage{latexsym}

\usepackage{url}
\usepackage{enumitem}
\usepackage{graphicx}
\usepackage{amssymb}
\usepackage{amsmath}
\usepackage{todonotes}
\usepackage{booktabs}
\usepackage{changepage}

\aclfinalcopy

\title{Deep Contextualized Word Embeddings in Transition-Based and
  Graph-Based Dependency Parsing -- A Tale of Two Parsers Revisited\thanks{~~We gratefully acknowledge the inspiration for our subtitle in the seminal paper by \newcite{zhang08emnlp}.}}

\author{Artur Kulmizev~~~Miryam de Lhoneux~~~Johannes Gontrum~~~Elena Fano~~~Joakim Nivre \\
  Department of Linguistics and Philology, Uppsala University \\
{\normalsize  \{\tt artur.kulmizev,miryam.de\_lhoneux,joakim.nivre\}@lingfil.uu.se }\\
{\normalsize  \{\tt johannes.gontrum.4608,elena.fano.3249\}@student.uu.se }\\}

\date{}

\begin{document}
\maketitle

\begin{abstract}
  Transition-based and graph-based dependency parsers have previously been shown 
  to have 
  complemen\-tary strengths and weaknesses: 
  transition-based parsers exploit rich structural features but
  suffer from error propagation, while graph-based parsers 
  benefit from global
  optimization but have restricted feature scope. 
  In this paper, we
  show that, even though some details of the picture have changed after the
  switch to neural networks and continuous representations, the basic
  trade-off between rich features and global optimization remains essentially the same. 
  Moreover, we show that deep contextualized word embeddings,
  which allow parsers to pack
  information about global sentence structure into local feature
  representations, benefit transition-based parsers more than graph-based
  parsers, making the two approaches virtually equivalent in terms of both
  accuracy and error profile. We argue that the reason is that these 
  representations help prevent search errors and thereby allow
  transition-based parsers to better exploit their inherent strength of making
  accurate local decisions. We support this explanation by an error
  analysis of parsing experiments on 13 languages.
\end{abstract}

\section{Introduction}
For more than a decade, research on data-driven dependency parsing has 
been dominated by two approaches: transition-based parsing and graph-based parsing \citep{mcdonald07emnlp,mcdonald11cl}. Transition-based parsing reduces the 
parsing task 
to 
scoring 
single parse actions and is often combined with local optimization and greedy search algorithms. Graph-based parsing decomposes parse trees into subgraphs and relies on global optimization and exhaustive (or at least non-greedy) search to find the best tree. These 
radically different approaches often lead to comparable parsing accuracy, but with distinct error profiles indicative of their respective strengths and weaknesses, as shown by \citet{mcdonald07emnlp,mcdonald11cl}. 

In recent years, dependency parsing, like most of NLP, has 
shifted from linear models and discrete features 
to neural networks and continuous representations. 
This has 
led to substantial accuracy improvements for both transition-based and graph-based parsers and raises the question whether their complementary strengths and weaknesses are still relevant. In this paper, we replicate the analysis of \citet{mcdonald07emnlp,mcdonald11cl} for neural 
parsers. In addition, we investigate the impact of deep contextualized word representations \citep{peters18,devlin19} for both types of parsers. 

Based on what we know about the strengths and weaknesses of the two approaches, we hypothesize that deep contextualized word representations will benefit transition-based parsing more than graph-based parsing. 
The reason is that these representations make information about global sentence structure
available locally, thereby helping to prevent search errors in greedy transition-based parsing. 
The hypothesis is corroborated in experiments on 13 languages, and the error analysis 
supports our suggested explanation. We also find that deep contextualized word representations improve parsing accuracy for longer sentences, both for transition-based and graph-based parsers.

\section{Two Models of Dependency Parsing}
\label{sec:parsing}

After playing a marginal role in NLP for many years, dependency-based approaches to syntactic parsing have become mainstream during the last fifteen years. This is especially true if we consider languages other than English, ever since the influential CoNLL shared tasks on dependency parsing in 2006 \citep{buchholz06} and 2007 \citep{nivre07conll} with data from 19 languages. 

The transition-based approach to dependency parsing was pioneered by \citet{yamada03} and \citet{nivre03iwpt}, with inspiration from history-based parsing \citep{black92} and data-driven shift-reduce parsing \citep{veenstra00}. The
idea is to reduce the complex parsing task to the simpler task of predicting the next parsing action and to implement parsing as greedy search for the optimal sequence of actions, guided by a simple classifier trained on local parser configurations. This produces parsers that are very efficient, often with linear time complexity, and which can benefit from rich non-local features defined over parser configurations 
but which may suffer from compounding search errors. 

The graph-based approach to dependency parsing was developed by \citet{mcdonald05acl,mcdonald05emnlp}, building on earlier work by \citet{eisner96coling}. 
The
idea is to score dependency trees by a linear combination of scores of local subgraphs, often single 
arcs, and to implement parsing as exact search for the highest scoring tree under a globally optimized model. These parsers do not suffer from search errors but parsing algorithms are more complex and restrict the scope of features to local subgraphs.  

The terms \emph{transition-based} and \emph{graph-based} were coined by \citet{mcdonald07emnlp,mcdonald11cl}, who performed a contrastive error analysis of the two top-performing systems in the CoNLL 2006 shared task on multilingual dependency parsing: MaltParser \citep{nivre06conll} and  MSTParser \citep{mcdonald06conll}, which represented the state of the art in transition-based and graph-based parsing, respectively, at the time. Their analysis shows that, despite having almost exactly the same parsing accuracy when averaged over 13 languages, the two parsers have very distinctive error profiles. MaltParser is more accurate on short sentences, on short dependencies, on dependencies near the leaves of the tree, on nouns and prounouns, and on 
subject and object relations. MSTParser is more accurate on long sentences, on long dependencies, on dependencies near the root of the tree, on verbs,
and on coordination relations and sentence roots. 

\citet{mcdonald07emnlp,mcdonald11cl} argue that these patterns can be explained by the complementary strengths and weaknesses of the systems. The transition-based MaltParser prioritizes rich structural features, which enable accurate disambiguation in local contexts, but is limited by a locally optimized model and greedy algorithm, resulting in search errors for structures that require longer transition sequences. The graph-based MSTParser benefits from a globally optimized model and exact inference, which 
gives a better analysis of global sentence structure, but is more restricted in the features it can use, which limits its capacity to score local structures accurately.

\begin{figure}[tbp]
\centering
\includegraphics[width=\columnwidth]{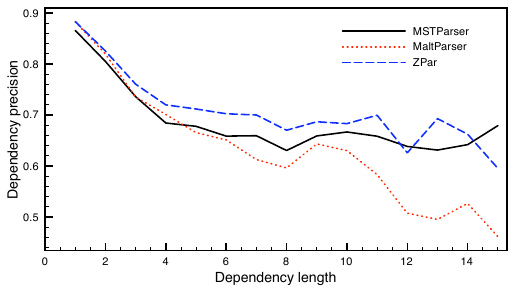}
\caption{Labeled precision by dependency length for MST (global--exhaustive--graph), Malt (local--greedy--transition) \index{and}and ZPar (global--beam--transition). From \citet{zhang12}.}
\label{fig:zhang12}
\end{figure}

Many of the developments in dependency parsing during the last decade can be understood in 
this light as attempts to mitigate the weaknesses of traditional transition-based and graph-based parsers without sacrificing 
their strengths. This may mean evolving the model structure through new transition systems \citep{nivre08cl,nivre09acl,kuhlmann11} or higher-order models for graph-based parsing \citep{mcdonald06eacl,carreras07,koo10acl}; it may mean 
exploring alternative learning strategies, in particular for transition-based parsing, where improvements have been achieved thanks to global structure learning \citep{zhang08emnlp,zhang11,andor16} and dynamic oracles \citep{goldberg12coling,goldberg13tacl}; it may mean 
using alternative search strategies, such as transition-based parsing with beam search \citep{johansson07,titov07iwpt,zhang08emnlp} or exact search \citep{huang10,kuhlmann11} or graph-based parsing with heuristic search to cope with the complexity of higher-order models, especially for non-projective parsing \citep{mcdonald06eacl,koo10emnlp,zhang12mcdonald}; or it may mean 
hybrid or ensemble systems \citep{sagae06naacl,nivre08acl,zhang08emnlp,bohnet12eacl}. 
A nice illustration of the impact of new techniques can be found in \citet{zhang12}, where an error analysis along the lines of \citet{mcdonald07emnlp,mcdonald11cl} shows that a transition-based parser using global learning and beam search (instead of local learning and greedy search) performs on par with graph-based parsers for long dependencies, while retaining the advantage of the original transition-based parsers on short dependencies (see Figure~\ref{fig:zhang12}).

Neural networks for dependency parsing, first explored by \citet{titov07iwpt} and \citet{attardi09}, have come to dominate the field during the last five years. While this has dramatically changed learning architectures and feature representations, most parsing models are still either transition-based \citep{chen14,dyer15,weiss15,andor16,kiperwasser16} or graph-based \citep{kiperwasser16,dozat17iclr}.
However, more accurate feature learning using continuous representations and nonlinear models has allowed parsing architectures to be simplified. Thus, 
most recent transition-based parsers have moved back to local learning and greedy inference, seemingly without losing accurracy \citep{chen14,dyer15,kiperwasser16}. Similarly, graph-based parsers again rely on first-order models and obtain no improvements from using higher-order models \citep{kiperwasser16,dozat17iclr}.

The increasing use of neural networks has also led to a convergence in feature representations and learning algorithms for transition-based and graph-based parsers. In particular, most recent systems rely on an encoder, typically in the form of a BiLSTM, that provides contextualized representations of the input words as input to the scoring of transitions -- in transition-based parsers -- or of dependency arcs -- in graph-based parsers. By making information about the \emph{global} sentence context available in \emph{local} word representations, this encoder can be assumed to mitigate error propagation for transition-based parsers and to widen the feature scope beyond individual word pairs for graph-based parsers. For both types of parsers, this also obviates the need for complex structural feature templates, as recently shown by \newcite{falenska19}. 
We should therefore expect neural transition-based and graph-based parsers to be not only more accurate than their non-neural counterparts but also more similar to each other in their error profiles. 

\section{Deep Contextualized Word Representations}

Neural parsers rely on vector representations of words as their primary input,
often in the form of pretrained word embeddings such as word2vec
\citep{mikolov13}, GloVe \citep{pennington14}, or fastText \citep{bojanowski16},
which are sometimes extended with character-based representations produced by recurrent
neural networks \citep{ballesteros15}.
These techniques 
assign a single static representation to each word type and therefore cannot capture context-dependent variation in meaning and syntactic behavior. 

By contrast, deep contextualized word representations encode words with respect to the sentential context in which they appear. 
Like word embeddings, such models are typically trained with a language-modeling objective, but yield sentence-level tensors as representations, instead of single vectors. 
These representations are typically produced by transferring a model's entire feature encoder --
be it a BiLSTM \cite{hochreiter1997long} or Transformer \cite{vaswani2017attention} -- to a target task, where the dimensionality of the tensor $\mathbf{S}$ is typically $\mathbf{S} \in \mathbb{R}^{N \times L \times D}$ for a sentence of length $N$, 
an encoder with $L$ layers, and word-level vectors of dimensionality $D$. The
advantage of such models, compared to the parser-internal encoders discussed in the previous section, is that they not only
produce contextualized representations but do so over several layers of abstraction, as captured by the model's different layers, and are pre-trained on corpora much larger than typical treebanks.

Deep contextualized embedding models have proven to be adept at a wide array of
NLP tasks, achieving state-of-the-art performance in standard Natural
Language Understanding (NLU) benchmarks, such as GLUE \cite{wang2019glue}.
Though many such models have been proposed, we adopt the two arguably
most popular ones for our experiments: ELMo and BERT. Both models have previously 
been used for dependency parsing \citep{che18,jawahar18,lim18,kondratyuk19,schuster19crosslingual}, but there has been no
systematic analysis of their impact on transition-based and graph-based parsers. 

\subsection{ELMo}
ELMo is a deep contextualized embedding model proposed by \citet{peters18},
which produces sentence-level representations yielded by a multi-layer BiLSTM
language model. ELMo is trained with a standard language-modeling objective,
in which a BiLSTM reads a sequence of $N$ learned context-independent embeddings
$\mathbf{w}_1, \ldots, \mathbf{w}_N$ (obtained via a character-level CNN) and produces a
context-dependent representation $\mathbf{h}_{j,k} =
\mathbf{BiLSTM}(\mathbf{w}_{1:N}, k)$, 
where $j$ $(1 \!\leq\! j \!\leq\! L)$ is the BiLSTM layer 
and $k$ is the index of the word
in the sequence. The output of the last layer $\mathbf{h}_{L,k}$ is then
employed in conjunction with a softmax layer to predict the next token at $k+1$.

The simplest way of transferring ELMo to a downstream task
is to encode the input sentence $S = w_1,\ldots,w_N$ by extracting the
representations from the BiLSTM at layer $L$ for each token $w_k \in S$: 
$\mathbf{h}_{L,1}, \ldots,\mathbf{h}_{L,N},$. However,
\citet{peters18} posit that the best way to take advantage of ELMo's representational
power is to compute a linear combination of BiLSTM layers:
\begin{equation}
  \label{eq:1}
  \mathbf{ELMo}_k = \gamma \sum_{j=0}^{L} s_{j} \mathbf{h}_{j,k}
\end{equation}
\noindent
where $s_j$ is a softmax-normalized task-specific parameter and $\gamma$ is a
task-specific scalar. \citet{peters18} demonstrate that this
scales the layers of linguistic abstraction encoded by the BiLSTM
for the task at hand.

\subsection{BERT}
BERT \cite{devlin19} is similar to ELMo in that it employs a
language-modeling objective over unannotated text in order to produce
deep contextualized embeddings. However, BERT differs from
ELMo in that, in place of a BiLSTM, it employs a bidirectional
Transformer \citep{vaswani2017attention}, which, among other factors, carries
the benefit of learning potential dependencies between words directly. This lies
in contrast to recurrent models, which may struggle to learn correspondences
between constituent signals when the time-lag between them is long
\cite{hochreiter2001gradient}.
For a token $w_{k}$ in sentence $S$, BERT's input representation is composed
by summing a word embedding $\mathbf{x}_{k}$, a position embedding $\mathbf{i}_{k}$, 
and a WordPiece embedding $\mathbf{s}_{k}$ \citep{wu2016google}:
$\mathbf{w}_{k} = \mathbf{x}_{k} + \mathbf{i}_{k} + \mathbf{s}_{k}$.

Each
  $\mathbf{w}_k \in S$ is passed to an $L$-layered BiTransformer, which is
  trained with a masked language modeling objective (i.e., randomly masking a
  percentage of input tokens and only predicting said tokens). For use in
  downstream tasks, \citet{devlin19} propose to extract the Transformer's
  encoding of each token $w_{k} \in S$ at layer $L$, which effectively produces $\mathbf{BERT}_k$.

\section{Hypotheses}

Based on our discussion in Section~\ref{sec:parsing}, we assume that transition-based and graph-based parsers still have distinctive error profiles due to the basic trade-off between rich structural features, which allow transition-based parsers to make accurate local decisions, and global learning and exact search, which give graph-based parsers an advantage with respect to global sentence structure. At the same time, we expect the differences to be less pronounced than they were ten years ago because of the convergence in neural architectures and feature representations. But how will the addition of deep contextualized word representations affect the behavior of the two parsers?  

Given recent recent work showing that deep contextualized word representations
incorporate rich information about syntactic structure
\citep{goldberg19,liu19,tenney19,hewitt19}, we hypothesize that transition-based
parsers have most to gain from these representations because it will improve
their capacity to make decisions informed by global sentence structure and
therefore reduce the number of search errors. Our main hypothesis can be stated
as follows:

\vspace{1.5mm}
\begin{adjustwidth}{1em}{1em}
    Deep contextualized word representations
are more effective\,at\,reducing\,errors
in transition-based parsing than in graph-based parsing.
\end{adjustwidth}
\vspace{1.5mm}
If this holds true, then the analysis of \citet{mcdonald07emnlp,mcdonald11cl} suggests that the differential error reduction should be especially visible on phenomena such as:
\begin{enumerate}[noitemsep,topsep=5pt]
\item longer dependencies,
\item dependencies closer to the root,
\item certain parts of speech,
\item certain dependency relations,
\item longer sentences.
\end{enumerate}
The error analysis will consider all these factors as well as non-projective dependencies.

\section{Experimental Setup}

\subsection{Parsing Architecture}
\label{sec:parser}
To be able to compare transition-based and graph-based parsers under equivalent conditions,
we use and extend UUParser\footnote{\url{https://github.com/UppsalaNLP/uuparser}} \citep{delhoneux17conll,smith18conll}, an evolution of bist-parser \citep{kiperwasser16}, 
which supports transition-based and graph-based parsing with a common infrastructure 
but different scoring models and parsing algorithms. 

For an input sentence $S = w_{1}, \ldots, w_{N}$, the parser creates a
sequence of vectors $\mathbf{w}_{1:N}$, where the vector $\mathbf{w}_k = \mathbf{x}_k \circ \textsc{BiLstm}(c_{1:M})$
representing input word $w_k$ is the concatenation of a pretrained word embedding $\mathbf{x}_k$
and a character-based
embedding $\textsc{BiLstm}(c_{1:M})$ obtained by running a \mbox{BiLSTM} over the
character sequence $c_{1:M}$ 
of $w_k$. 
Finally, each input element is represented by a \mbox{BiLSTM} vector, $\mathbf{h}_k = \textsc{BiLstm}(\mathbf{w}_{1:N},k)$.

In transition-based parsing, the \mbox{BiLSTM} vectors are input to a multi-layer perceptron (MLP) for scoring transitions, 
using the arc-hybrid transition system from \citet{kuhlmann11}
extended with a \textsc{Swap} transition to 
allow the construction of non-projective dependency trees \cite{nivre09acl,delhoneux17iwpt}.
The scoring is based on the top three words on the stack and the first word of the buffer, and the input to the MLP includes the BiLSTM vectors for these words as well as their leftmost and rightmost dependents (up to 12 words in total).

In graph-based parsing, the BiLSTM vectors are input to an MLP for scoring all possible dependency relations under an arc-factored model, meaning that only the vectors corresponding to the head and dependent are part of the input (2 words in total). The parser then extracts a maximum spanning tree over the score matrix using the Chu-Liu-Edmonds (CLE) algorithm\footnote{We use the implementation from \citet{qi2018universal}.} \citep{edmonds67} which allows us to construct non-projective trees.

It is important to note that, while we acknowledge the existence of
graph-based parsers that outperform the implementation of \citet{kiperwasser16}, 
such models do not meet our criteria for 
systematic comparison.
The parser by \citet{dozat17conll} is very similar, but employs the MLP as a further step in
the featurization process prior to scoring via a biaffine classifier. To keep the comparison 
as exact as possible, 
we forego comparing our transition-based systems to the
\citet{dozat17conll} parser (and its numerous modifications). 
In addition, preliminary experiments showed that our chosen graph-based parser outperforms
its transition-based counterpart, which was itself competitive in the CoNLL 2018
shared task \cite{zeman18conll}.

\subsection{Input Representations}

In our experiments, we evaluate three pairs of systems -- differing only in their input representations. 
The first is a baseline that represents tokens by $\mathbf{w}_k = \mathbf{x}_k \circ \textsc{BiLstm}(c_{1:M})$, as described in Section \ref{sec:parser}. The word embeddings $\mathbf{x}_k$ are initialized via
pretrained fastText vectors ($\mathbf{x}_k \in \mathbb{R}^{300}$)
\cite{grave2018learning}, which are updated for the parsing task. We term these
transition-based and graph-based baselines \textsc{tr} and \textsc{gr}.

For the ELMo
experiments, we make use of pretrained models
provided by \citet{che18}, who train ELMo on 20 million words
randomly sampled from raw WikiDump and Common Crawl datasets for 44
languages. We encode each gold-segmented sentence in our treebank via 
the ELMo model for that language, which yields a tensor
$\mathbf{S}_{\mathrm{ELMo}} = \mathbb{R}^{N \times L \times D}$, where $N$ is the number
of words in the sentence, $L = 3$ is the number of ELMo layers, and $D
= 1024$ is the ELMo vector dimensionality. Following \citet{peters18} (see Eq. \ref{eq:1}),
we learn a linear combination and a task-specific $\gamma$ of each token's ELMo representation,
which yields a vector $\mathbf{ELMo}_{k} \in \mathbb{R}^{1024}$. We then
concatenate this vector with $\mathbf{w}_k$ and
pass it to the \mbox{BiLSTM}. We call the transition-based and graph-based
systems enhanced with ELMo \textsc{tr+E} and
\textsc{gr+E}.

For the BERT
experiments, we employ the pretrained multilingual
cased model provided by Google,%
\footnote{\url{https://github.com/google-research/bert}}\!\!
\footnote{Except for Chinese, for which we make use
  of a separate, pretrained model.} 
which is trained on the concatenation of
WikiDumps for the top 104 languages with the largest Wikipedias.\footnote{See
  sorted list here: \url{https://meta.wikimedia.org/wiki/List_of_Wikipedias}} The model's
parameters feature a 12-layer transformer trained with 768 hidden units and 12
self-attention heads. In order to obtain a word-level vector for each token in a
sentence, we experimented with a variety of representations: namely,
concatenating each transformer layer's word representation into a single vector
$\mathbf{w}_{concat} \in \mathbb{R}^{768*12}$, employing the last layer's
representation, or learning a linear combination over a range of layers, as we
do with ELMo (e.g., via Eq. \ref{eq:1}). In a preliminary set of experiments, we found that the
latter approach over layers 4--8 consistently yielded the best results, and
thus chose to adopt this method going forward. Regarding tokenization, we select
the vector for the first subword token, as produced by the native BERT
tokenizer. Surprisingly, this gave us better results than averaging subword
token vectors in a preliminary round of experiments. Like with the ELMo
representations, we concatenate each BERT vector $\mathbf{BERT}_{k} \in
\mathbb{R}^{768}$ with $\mathbf{w}_k$ and pass it to the respective
\textsc{tr+B} and \textsc{gr+B} parsers.

It is important to note that while the ELMo models we work with are
\textit{monolingual}, the BERT model is \textit{multilingual}.
In other words, while the standalone ELMo models were trained on the tokenized WikiDump
and CommonCrawl for each language respectively, the BERT model was trained only on
the former, albeit simultaneously for 104 languages. 
This means that the models are not strictly comparable, and it is an interesting
question whether either of the models has an advantage in terms of training regime. 
However, since our purpose is not to compare the two models but
to study their impact on parsing, we leave this question 
for future work.

\subsection{Language and Treebank Selection}

\begin{table}[t!]
\centering
\renewcommand{\tabcolsep}{3pt}
\begin{tabular}{@{}llllr@{}}
  \textbf{Language} & \textbf{Treebank}  & \textbf{Family} & \textbf{Order} & \textbf{Train} \\ \hline
Arabic   & PADT    & non-IE   & VSO    & 6.1k       \\
Basque   & BDT    & non-IE       & SOV  & 5.4k       \\
Chinese  & GSD       & non-IE    & SVO    & 4.0k     \\
English  & EWT    & IE       & SVO   & 12.5k    \\
Finnish  & TDT        & non-IE & SVO    & 12.2k    \\
Hebrew   & HTB     & non-IE      & SVO     & 5.2k    \\
Hindi    & HDTB     & IE    & SOV     & 13.3k    \\
Italian  & ISDT    & IE       & SVO   & 13.1k    \\
Japanese & GSD      & non-IE     & SOV     & 7.1k    \\
Korean   & GSD      & non-IE     & SOV     & 4.4k    \\
Russian  & SynTagRus  & IE  & SVO     & 48.8k    \\
Swedish  & Talbanken    & IE   & SVO   & 4.3k    \\
Turkish  & IMST     & non-IE    & SOV      & 3.7k   \\ 
\end{tabular}
\caption{Languages and treebanks used in experiments. Family = Indo-European (IE) or not. Order = dominant word order according to WALS \citep{haspelmath05}. Train = number of training sentences.}
\label{tab:languages}
\end{table}

For treebank selection, we rely on the criteria proposed by 
\citet{delhoneux2017old} and adapted by \citet{smith18emnlp} to have languages from different language families, with different morphological complexity, different scripts and character set sizes, different training sizes and domains, and with good annotation quality.  This gives us 13 treebanks from UD v2.3 \cite{ud23}, information about which is shown in Table ~\ref{tab:languages}.

\subsection{Parser Training and Evaluation}

In all experiments, we train parsers with default settings\footnote{All
  hyperparameters are specified in the supplementary material (Part A).} for 30 epochs and select the model with the best labeled attachment score on the dev set. For each combination of model and training set, we repeat this procedure three times with different random seeds, apply the three selected models to the test set, and report the average result. 

\subsection{Error Analysis}

In order to conduct an error analysis along the lines of \citet{mcdonald07emnlp,mcdonald11cl}, 
we extract all sentences from the smallest development set in our treebank sample 
(Hebrew HTB, 484 sentences) and sample the same number of sentences from each of 
the other development sets (6,292 sentences in total). For each system, we then extract
parses of these sentences for the three training runs with different random seeds 
(18,876 predictions in total). Although it could be interesting to look at each language separately, 
we follow \citet{mcdonald07emnlp,mcdonald11cl} and base our main analysis on all languages 
together to prevent data sparsity for longer dependencies, longer sentences, etc.%
\footnote{The supplementary material contains tables for the error analysis (Part B) and graphs for each language (Part C).}

\section{Results and Discussion}
\label{sec:results}

Table~\ref{tab:results} shows labeled attachment scores for the six parsers on
all languages, averaged over three training runs with random seeds.
The results clearly corroborate our main hypothesis. While ELMo and BERT provide
significant improvements for both transition-based and graph-based parsers, the
magnitude of the improvement is greater in the transition-based case: 3.99
vs.\ 2.85 for ELMo and 4.47 vs. 3.13 for BERT. In terms of error reduction, this
corresponds to 21.1\% vs.\ 16.5\% for ELMo and 22.5\% vs.\ 17.4\% for BERT.
The differences in error reduction are statistically significant at $\alpha = 0.01$ (Wilcoxon).

Although both parsing accuracy and absolute improvements vary across languages, the overall
trend is remarkably consistent and the transition-based parser improves more with both ELMo and BERT
for every single language. Furthermore, a linear mixed effect model analysis
reveals that, when accounting for language as a random effect, there are no
significant interactions between the improvement of each model (over its
respective baseline) and factors such as language family (IE vs.\ non-IE),
dominant word order, or number of training sentences. In other words, the improvements for both parsers  seem to be largely independent of treebank-specific factors. Let us now see to what extent they can be explained by the error analysis.

\begin{table}[t!]
\centering
\renewcommand{\tabcolsep}{1.4pt}
\begin{tabular}{lrrrrrr}
\multicolumn{1}{l}{\textbf{Language}} & 
\multicolumn{1}{r}{\textsc{tr}} & 
\multicolumn{1}{r}{\textsc{gr}} & 
\multicolumn{1}{r}{\textsc{tr+E}} & 
\multicolumn{1}{r}{\textsc{gr+E}} & 
\multicolumn{1}{r}{\textsc{tr+B}} & 
\multicolumn{1}{r}{\textsc{gr+B}} \\
\hline
Arabic & 79.1 & ~~~79.9 & 82.0 & 81.7 & 81.9 & 81.8 \\
Basque & 73.6 & 77.6 & 80.1 & 81.4 & 77.9 & 79.8 \\
Chinese & 75.3 & 76.7 & 79.8 & 80.4 & 83.7 & 83.4 \\
English & 82.7 & 83.3 & 87.0 & 86.5 & 87.8 & 87.6 \\
Finnish & 80.0 & 81.4 & 87.0 & 86.6 & 85.1 & 83.9 \\
Hebrew & 81.1 & 82.4 & 85.2 & 85.9 & 85.5 & 85.9 \\
Hindi & 88.4 & 89.6 & 91.0 & 91.2 & 89.5 & 90.8 \\
Italian & 88.0 & 88.2 & 90.9 & 90.6 & 92.0 & 91.7 \\
Japanese & 92.1 & 92.2 & 93.1 & 93.0 & 92.9 & 92.1 \\
Korean & 79.6 & 81.2 & 82.3 & 82.3 & 83.7 & 84.2 \\
Russian & 88.3 & 88.0 & 90.7 & 90.6 & 91.5 & 91.0 \\
Swedish & 80.5 & 81.6 & 86.9 & 86.2 & 87.6 & 86.9 \\
Turkish & 57.8 & 61.2 & 62.6 & 63.8 & 64.2 & 64.9 \\
  \hline
\textbf{Average} & \textbf{80.5} & \textbf{81.8} & \textbf{84.5} & \textbf{84.6} & \textbf{84.9} & \textbf{84.9} \\ 
\end{tabular}
\caption{Labeled attachment score on 13 languages for parsing models with and without deep contextualized word representations.}
\label{tab:results}
\end{table}

\begin{figure}[t!]
\centering
\includegraphics[width=\columnwidth]{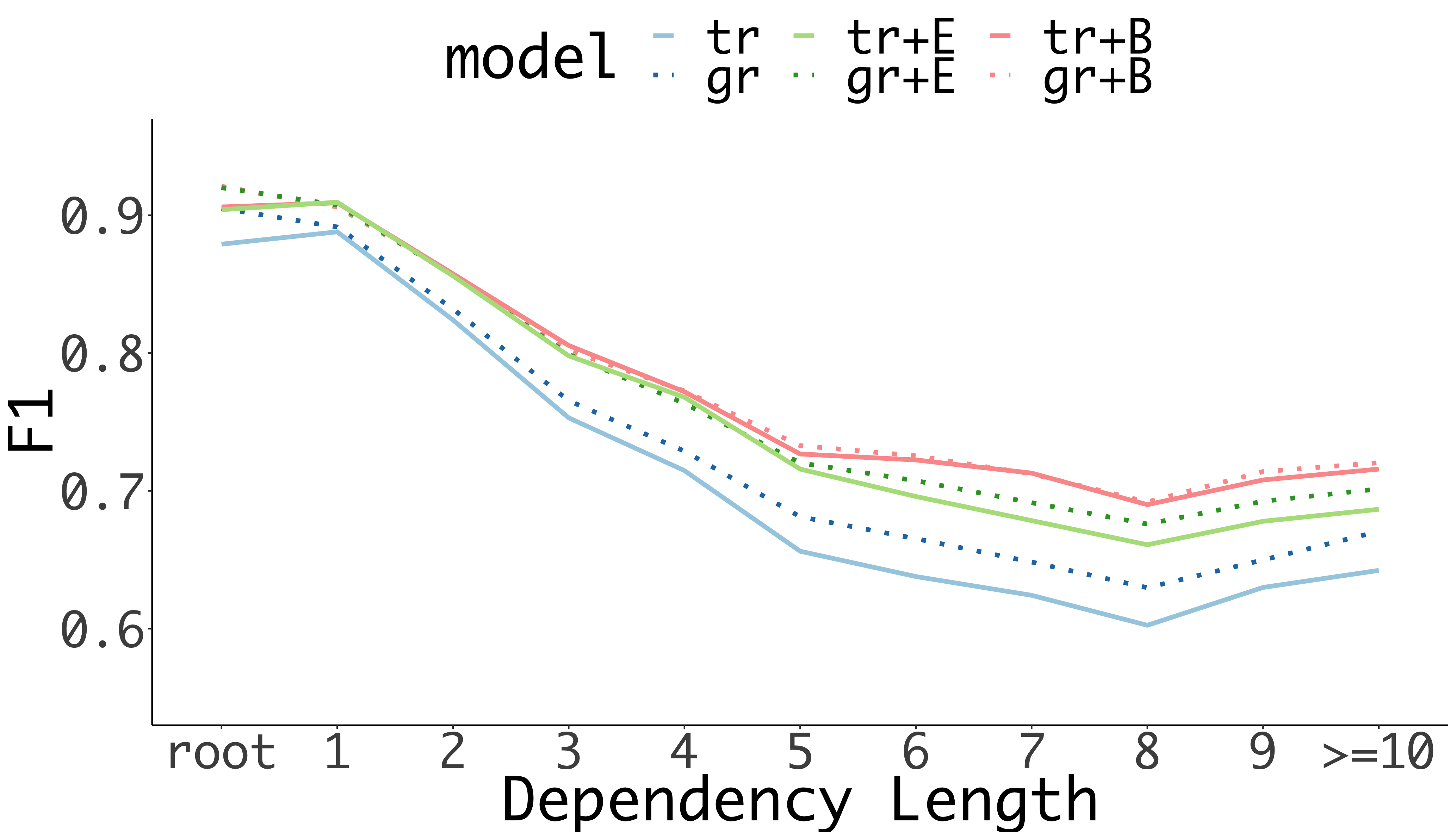}
\caption{Labeled F-score by dependency length.}
\label{fig:deplen}
\end{figure}

\begin{figure}[t!]
\centering
\includegraphics[width=\columnwidth]{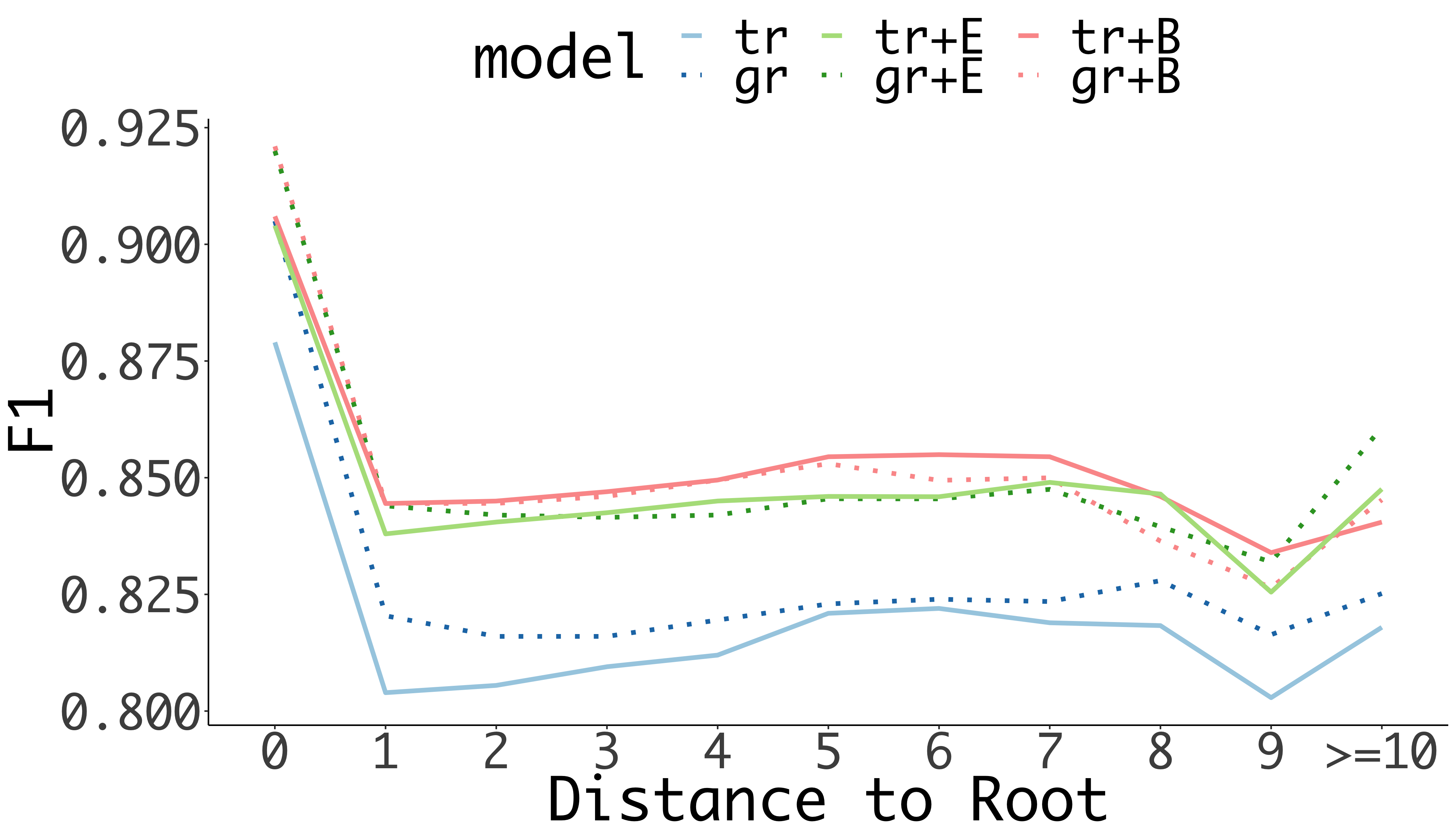}
\caption{Labeled F-score by distance to root.}
\label{fig:rootdist}
\end{figure}

\begin{figure}[t!]
\centering
\includegraphics[width=\columnwidth]{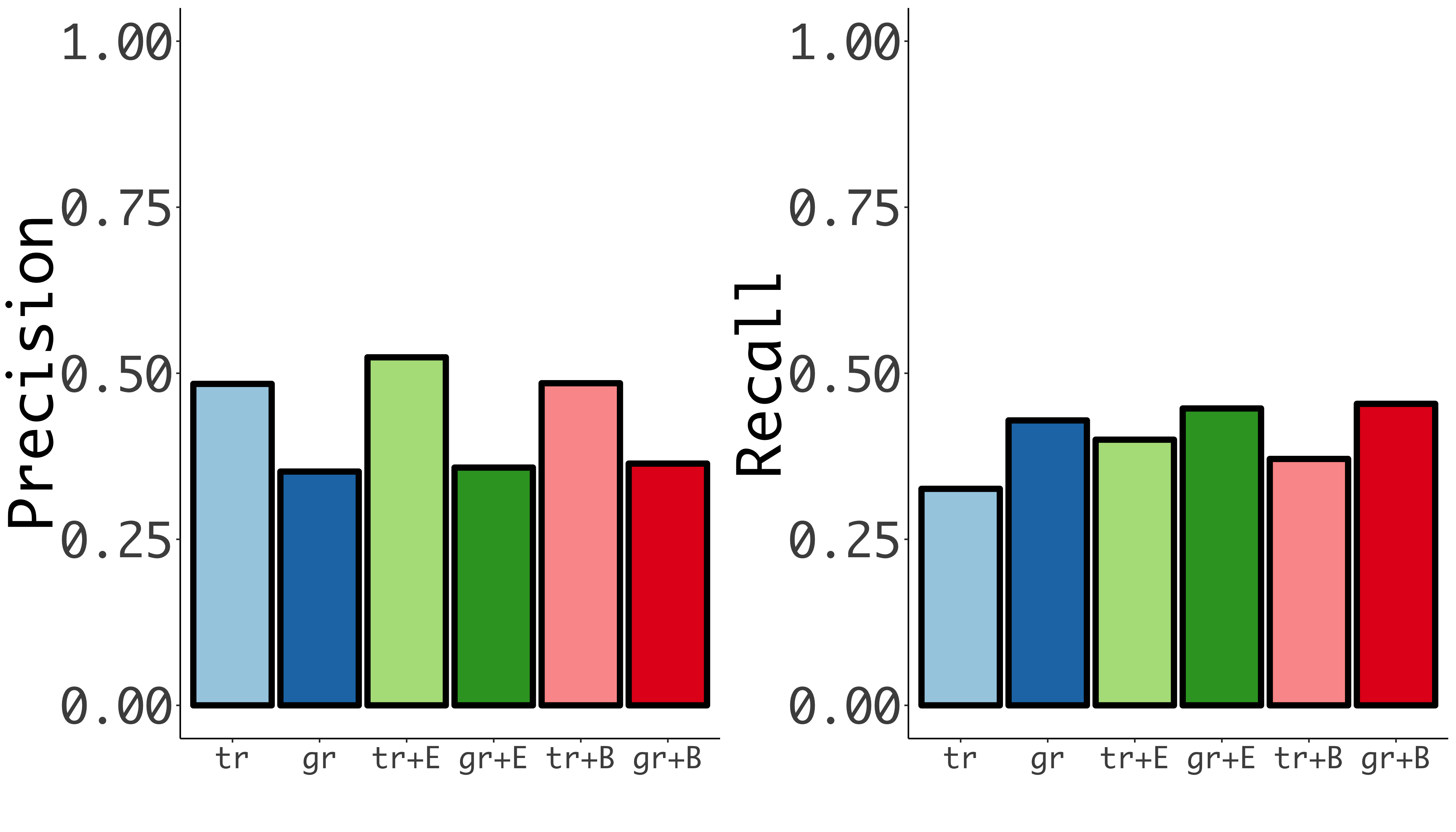}
\caption{Labeled precision (left) and recall (right) for non-projective dependencies.}
\label{fig:proj}
\end{figure}

\subsection{Dependency Length}

Figure \ref{fig:deplen} shows labeled F-score for dependencies of
different lengths, where the length of a dependency between words $w_{i}$ and $w_{j}$ is 
equal to $| i - j |$ (and with root tokens in a special bin on the far left). 
For the baseline parsers, we see that the curves diverge with increasing length,
clearly indicating that the transition-based parser still suffers from search errors on
long dependencies, which require longer transition sequences for their construction.
However, the differences are much 
smaller than in \citet{mcdonald07emnlp,mcdonald11cl}
and the transition-based parser no longer has an advantage for short dependencies, which
is consistent with the BiLSTM architecture providing the parsers with more similar features
that help the graph-based parser overcome the limited scope of the first-order model.

Adding deep contextualized word representations clearly helps the
transition-based parser to perform better on longer dependencies. For ELMo there
is still a discernible difference for dependencies longer than 5, but for BERT
the two curves are almost indistinguishable throughout the whole range.
This could be related to the aforementioned intuition that a Transformer captures 
long dependencies more effectively than a BiLSTM (see \citet{tran2018importance}
for contrary observations, albeit for different tasks).
The overall trends for both baseline and enhanced models are quite 
consistent across languages, although with large variations in accuracy levels.

\subsection{Distance to Root}

Figure \ref{fig:rootdist} reports labeled F-score for dependencies
at different distances from the root of the tree, where distance is measured by
the number of arcs in the path from the root. 
There is a fairly strong (inverse) correlation between dependency length and distance to the root, 
so it is not surprising that the plots in Figure \ref{fig:rootdist} largely show the mirror image of the plots in Figure \ref{fig:deplen}. For the baseline parsers, the graph-based parser has a clear advantage for dependencies near the root (including the root itself), but the transition-based parser closes the gap with increasing distance.\footnote{At the very end, the curves appear to diverge again, but the data is very sparse in this part of the plot.} For ELMo and BERT, the curves are much more similar, with only a slight advantage for the graph-based parser near the root and with the transition-based BERT parser being superior from distance 5 upwards. The main trends are again similar across all languages.

\subsection{Non-Projective Dependencies}

Figure~\ref{fig:proj} shows precision and recall specifically for non-projective dependencies.
We see that there is a clear tendency for the transition-based parser to have better precision 
and the graph-based parser better recall.\footnote{Incidentally, the same pattern is reported by
  \citet{mcdonald07emnlp,mcdonald11cl}, even though the techniques for
  processing non-projective dependencies are different in that study:
  pseudo-projective parsing \citep{nivre05acl} for the transition-based parser
  and approximate second-order non-projective parsing \citep{mcdonald06eacl} for
  the graph-based parser.} In other words, non-projective dependencies are more
likely to be correct when they are predicted by the transition-based parser
using the swap transition, but real non-projective dependencies are more likely
to be found by the graph-based parser using a spanning tree algorithm.
Interestingly, adding deep contextualized word representations has almost no effect on
the graph-based parser,\footnote{The breakdown per language shows 
marginal improvements for the enhanced graph-based models on a few languages, canceled
out by equally marginal degradations on others.} 
while especially the ELMo embeddings improve both precision
and recall for the transition-based parser.

\subsection{Parts of Speech and Dependency Types}

Thanks to the cross-linguistically consistent UD annotations, we can relate
errors to linguistic cate\-gories more systematically than in the old study. The
main impression, however, is that there are very few clear differences, which is
again indicative of the convergence between the two parsing approaches. We
highlight the most notable differences and refer to the supplementary material 
(Part B) for the full results.

Looking first at parts of speech, the baseline graph-based parser is slightly
more accurate on verbs and nouns than its transition-based counterpart, which is
consistent with the old study for verbs but not for nouns. After adding the deep
contextualized word representations, both differences are essentially
eliminated.

With regard to dependency relations, the baseline graph-based parser has better
precision and recall than the baseline transition-based parser for the relations
of coordination (conj), which is consistent with the old study, as well as
clausal subjects (csubj) and clausal complements (ccomp), which are relations
that involve verbs in clausal structures. Again, the differences are greatly
reduced in the enhanced parsing models, especially for clausal complements,
where the transition-based parser with ELMo representations is even slightly
more accurate than the graph-based parser.

\begin{figure}[t!]
\centering
\includegraphics[width=\columnwidth]{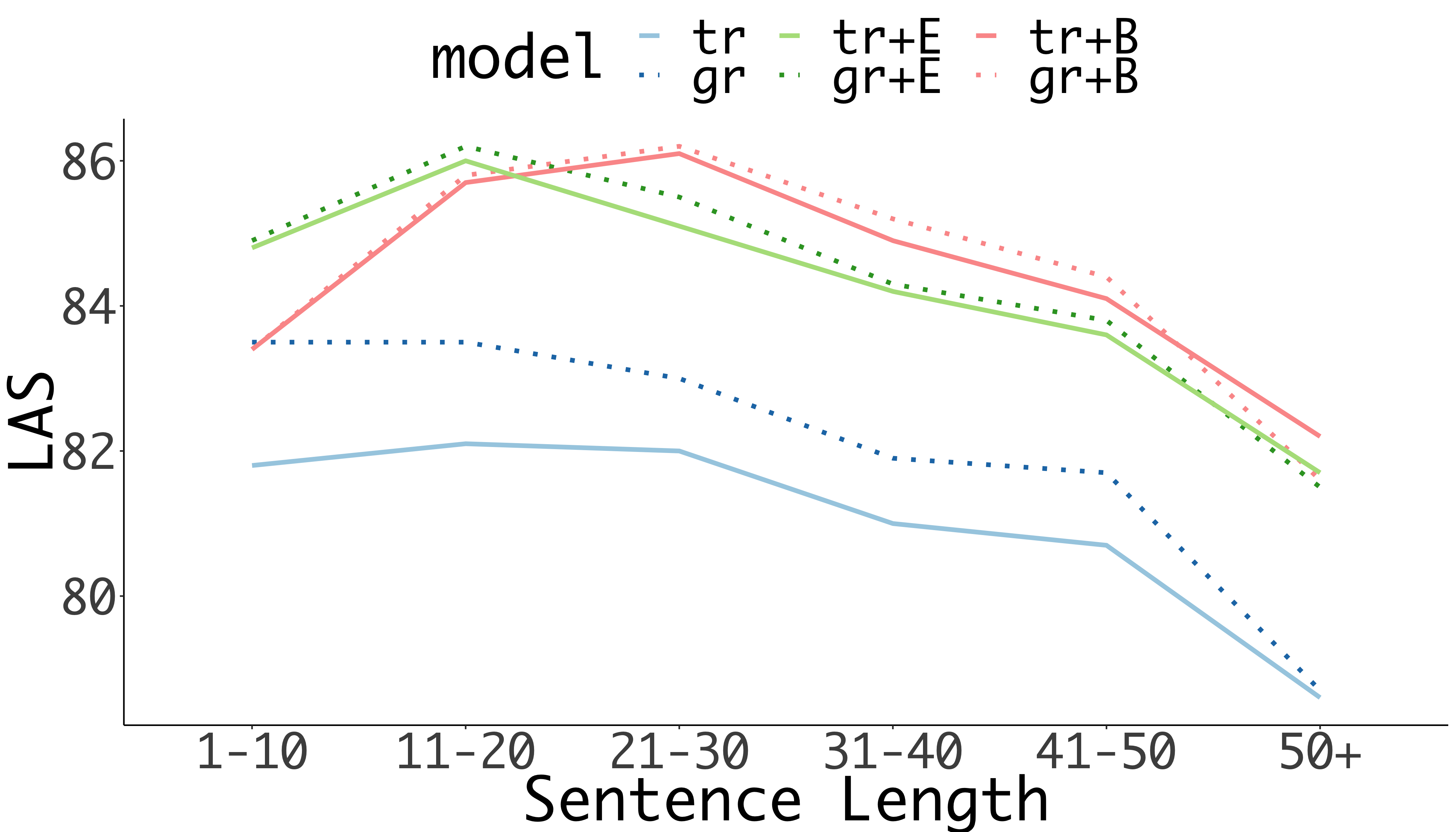}
\caption{Labeled attachment score by sentence length.}
\label{fig:senlen}
\end{figure}

\subsection{Sentence Length}

Figure \ref{fig:senlen} plots labeled attachment score for sentences of different 
lengths, measured by number of words in bins of 1--10, 11--20, etc. Here we find the
most unexpected results of the study. First of all, although the baseline parsers exhibit
the familiar pattern of accuracy decreasing with sentence length, it is not the transition-based 
but the graph-based parser that is more accurate on short sentences and degrades faster.
In other words, although the transition-based parser still seems to suffer from search errors,
as shown by the results on dependency length and distance to the root, it no longer seems to
suffer from error propagation in the sense that earlier errors make later errors more probable. 
The most likely explanation for this is the improved training for transition-based parsers 
using dynamic oracles and aggressive exploration to learn how to behave optimally also in non-optimal configurations \citep{goldberg12coling,goldberg13tacl,kiperwasser16}. 

Turning to the models with deep contextualized word representations, we find
that transition-based and graph-based parsers behave more similarly, which
is in line with our hypotheses. However, the most noteworthy result is that
accuracy \emph{improves} with increasing sentence
length. For ELMo this holds only from 1--10 to 11--20, but for BERT it holds up
to 21--30, and even sentences of length 31--40 are parsed with higher accuracy
than sentences of length 1--10. 
A closer look at the breakdown per language reveals that this picture is slightly 
distorted by different sentence length distributions in different languages.
More precisely, high-accuracy languages seem to have a higher proportion 
of sentences of mid-range length, causing a slight boost in the accuracy scores
of these bins, and no single language exhibits exactly the patterns shown in Figure~\ref{fig:senlen}.
Nevertheless, several languages exhibit an increase in accuracy from the first to the second bin
or from the second to the third bin for one or more of the enhanced models (especially the BERT models). 
And almost all languages show a less steep degradation for the enhanced models, 
clearly indicating that deep contextualized word representations improve the capacity to parse longer sentences.

\section{Conclusion}
In this paper, we have essentially replicated the study of
\citet{mcdonald07emnlp,mcdonald11cl} for neural parsers. In the baseline
setting, where parsers use pre-trained word embeddings and character
representations fed through a BiLSTM, we can still discern the basic trade-off
identified in the old study, with the transition-based parser
suffering from search errors leading to lower accuracy on 
long dependencies and dependencies near the root of the tree. However,
important details of the picture have changed. The graph-based parser is
now as accurate as the transition-based parser on shorter dependencies and
dependencies near the leaves of the tree, thanks to improved representation
learning that overcomes the limited feature scope of the first order model. 
And with respect to sentence length, the pattern has actually been reversed,
with the graph-based parser being more accurate on short sentences and the
transition-based parser gradually catching up thanks to new training methods that
prevent error propagation.

When adding deep contextualized word representations, the behavior of the two
parsers converge even more, and the transition-based parser in particular
improves with respect to longer dependencies and dependencies near the root,
as a result of fewer search errors thanks to enhanced information
about the global sentence structure. One of the most striking results, however,
is that both parsers improve their accuracy on longer sentences, with some 
models for some languages in fact being more accurate on medium-length
sentences than on shorter sentences. This is a milestone in parsing research,
and more research is needed to explain it.

In a broader perspective, we hope that future studies on dependency parsing will 
take the results obtained here into account
and extend them by investigating other parsing approaches and neural network architectures.
Indeed, given the 
rapid development of new representations and architectures, future work should 
include analyses of how all components in neural parsing architectures (embeddings, 
encoders, decoders) contribute to distinct error profiles (or lack thereof).

\section*{Acknowledgments}

We want to thank Ali Basirat, Christian Hardmeier, Jamie Henderson, Ryan McDonald, Paola Merlo, Gongbo Tang, and the EMNLP reviewers and area chairs for valuable feedback on preliminary versions of this paper.
We acknowledge the computational resources provided by CSC in Helsinki and Sigma2 in Oslo through NeIC-NLPL (www.nlpl.eu). 

\bibliographystyle{acl_natbib}
\bibliography{expanded,main}

\begin{thebibliography}{74}
\expandafter\ifx\csname natexlab\endcsname\relax\def\natexlab#1{#1}\fi

\bibitem[{Andor et~al.(2016)Andor, Alberti, Weiss, Severyn, Presta, Ganchev,
  Petrov, and Collins}]{andor16}
Daniel Andor, Chris Alberti, David Weiss, Aliaksei Severyn, Alessandro Presta,
  Kuzman Ganchev, Slav Petrov, and Michael Collins. 2016.
\newblock Globally normalized transition-based neural networks.
\newblock In \emph{Proceedings of the 54th Annual Meeting of the Association
  for Computational Linguistics ({ACL})}, pages 2442--2452.

\bibitem[{Attardi et~al.(2009)Attardi, Dell’Orletta, Simi, and
  Turian}]{attardi09}
Giuseppe Attardi, Felice Dell’Orletta, Maria Simi, and Joseph Turian. 2009.
\newblock Accurate dependency parsing with a stacked multilayer perceptron.
\newblock In \emph{Proceedings of EVALITA 2009}.

\bibitem[{Ballesteros et~al.(2015)Ballesteros, Dyer, and Smith}]{ballesteros15}
Miguel Ballesteros, Chris Dyer, and Noah~A. Smith. 2015.
\newblock Improved transition-based parsing by modeling characters instead of
  words with {LSTM}s.
\newblock In \emph{Proceedings of the Conference on Empirical Methods in
  Natural Language Processing ({EMNLP})}, pages 349--359.

\bibitem[{Black et~al.(1992)Black, Jelinek, Lafferty, Magerman, Mercer, and
  Roukos}]{black92}
Ezra Black, Frederick Jelinek, John~D. Lafferty, David~M. Magerman, Robert~L.
  Mercer, and Salim Roukos. 1992.
\newblock Towards history-based grammars: Using richer models for probabilistic
  parsing.
\newblock In \emph{Proceedings of the 5th DARPA Speech and Natural Language
  Workshop}, pages 31--37.

\bibitem[{Bohnet and Kuhn(2012)}]{bohnet12eacl}
Bernd Bohnet and Jonas Kuhn. 2012.
\newblock The best of both worlds -- a graph-based completion model for
  transition-based parsers.
\newblock In \emph{Proceedings of the 13th Conference of the European Chpater
  of the Association for Computational Linguistics ({EACL})}, pages 77--87.

\bibitem[{Bojanowski et~al.(2016)Bojanowski, Grave, Joulin, and
  Mikolov}]{bojanowski16}
Piotr Bojanowski, Edouard Grave, Armand Joulin, and Tomas Mikolov. 2016.
\newblock Enriching word vectors with subword information.
\newblock \emph{arXiv preprint arXiv:1607.04606}.

\bibitem[{Buchholz and Marsi(2006)}]{buchholz06}
Sabine Buchholz and Erwin Marsi. 2006.
\newblock {CoNLL-X} shared task on multilingual dependency parsing.
\newblock In \emph{Proceedings of the 10th Conference on Computational Natural
  Language Learning ({CoNLL})}, pages 149--164.

\bibitem[{Carreras(2007)}]{carreras07}
Xavier Carreras. 2007.
\newblock Experiments with a higher-order projective dependency parser.
\newblock In \emph{Proceedings of the CoNLL Shared Task of EMNLP-CoNLL 2007},
  pages 957--961.

\bibitem[{Che et~al.(2018)Che, Liu, Wang, Zheng, and Liu}]{che18}
Wanxiang Che, Yijia Liu, Yuxuan Wang, Bo~Zheng, and Ting Liu. 2018.
\newblock Towards better {UD} parsing: Deep contextualized word embeddings,
  ensemble, and treebank concatenation.
\newblock In \emph{Proceedings of the {C}o{NLL} 2018 Shared Task: Multilingual
  Parsing from Raw Text to Universal Dependencies}, pages 55--64.

\bibitem[{Chen and Manning(2014)}]{chen14}
Danqi Chen and Christopher Manning. 2014.
\newblock A fast and accurate dependency parser using neural networks.
\newblock In \emph{Proceedings of the Conference on Empirical Methods in
  Natural Language Processing ({EMNLP})}, pages 740--750.

\bibitem[{Devlin et~al.(2019)Devlin, Chang, Lee, and Toutanova}]{devlin19}
Jacob Devlin, Ming-Wei Chang, Kenton Lee, and Kristina Toutanova. 2019.
\newblock {BERT:} pre-training of deep bidirectional transformers for language
  understanding.
\newblock In \emph{Proceedings of the 2019 Conference of the North {A}merican
  Chapter of the Association for Computational Linguistics: Human Language
  Technologies}.

\bibitem[{Dozat and Manning(2017)}]{dozat17iclr}
Timothy Dozat and Christopher~D. Manning. 2017.
\newblock Deep biaffine attention for neural dependency parsing.
\newblock In \emph{Proceedings of the 5th International Conference on Learning
  Representations}.

\bibitem[{Dozat et~al.(2017)Dozat, Qi, and Manning}]{dozat17conll}
Timothy Dozat, Peng Qi, and Christopher~D. Manning. 2017.
\newblock Stanford's graph-based neural dependency parser at the conll 2017
  shared task.
\newblock In \emph{Proceedings of the CoNLL 2017 Shared Task: Multilingual
  Parsing from Raw Text to Universal Dependencies}, pages 20--30.

\bibitem[{Dyer et~al.(2015)Dyer, Ballesteros, Ling, Matthews, and
  Smith}]{dyer15}
Chris Dyer, Miguel Ballesteros, Wang Ling, Austin Matthews, and Noah~A. Smith.
  2015.
\newblock Transition-based dependency parsing with stack long short-term
  memory.
\newblock In \emph{Proceedings of the 53rd Annual Meeting of the Association
  for Computational Linguistics ({ACL})}, pages 334--343.

\bibitem[{Edmonds(1967)}]{edmonds67}
Jack Edmonds. 1967.
\newblock Optimum branchings.
\newblock \emph{Journal of Research of the National Bureau of Standards},
  71B:233--240.

\bibitem[{Eisner(1996)}]{eisner96coling}
Jason~M. Eisner. 1996.
\newblock Three new probabilistic models for dependency parsing: An
  exploration.
\newblock In \emph{Proceedings of the 16th International Conference on
  Computational Linguistics ({COLING})}, pages 340--345.

\bibitem[{Falenska and Kuhn(2019)}]{falenska19}
Agnieszka Falenska and Jonas Kuhn. 2019.
\newblock The (non-)utility of structural features in {B}i{LSTM}-based
  dependency parsers.
\newblock In \emph{Proceedings of the 57th Annual Meeting of the Association
  for Computational Linguistics ({ACL})}, pages 117--128.

\bibitem[{Goldberg(2019)}]{goldberg19}
Yoav Goldberg. 2019.
\newblock Assessing {BERT}'s syntactic abilities.
\newblock \emph{CoRR}, abs/1901.05287.

\bibitem[{Goldberg and Nivre(2012)}]{goldberg12coling}
Yoav Goldberg and Joakim Nivre. 2012.
\newblock A dynamic oracle for arc-eager dependency parsing.
\newblock In \emph{Proceedings of the 24th International Conference on
  Computational Linguistics ({COLING})}, pages 959--976.

\bibitem[{Goldberg and Nivre(2013)}]{goldberg13tacl}
Yoav Goldberg and Joakim Nivre. 2013.
\newblock Training deterministic parsers with non-deterministic oracles.
\newblock \emph{Transactions of the Association for Computational Linguistics},
  1:403--414.

\bibitem[{Grave et~al.(2018)Grave, Bojanowski, Gupta, Joulin, and
  Mikolov}]{grave2018learning}
Edouard Grave, Piotr Bojanowski, Prakhar Gupta, Armanpd Joulin, and Tomas
  Mikolov. 2018.
\newblock Learning word vectors for 157 languages.
\newblock In \emph{Proceedings of the International Conference on Language
  Resources and Evaluation (LREC 2018)}.

\bibitem[{Haspelmath et~al.(2005)Haspelmath, Dryer, Gil, and
  Comrie}]{haspelmath05}
Martin Haspelmath, Matthew~S. Dryer, David Gil, and Bernard Comrie. 2005.
\newblock \emph{Thw World Atlas of Language Structures}.
\newblock Oxford University Press.

\bibitem[{Hewitt and Manning(2019)}]{hewitt19}
John Hewitt and Christopher~D. Manning. 2019.
\newblock A structural probe for finding syntax in word representations.
\newblock In \emph{Proceedings of the 2019 Conference of the North {A}merican
  Chapter of the Association for Computational Linguistics: Human Language
  Technologies}.

\bibitem[{Hochreiter et~al.(2001)Hochreiter, Bengio, Frasconi, Schmidhuber
  et~al.}]{hochreiter2001gradient}
Sepp Hochreiter, Yoshua Bengio, Paolo Frasconi, J{\"u}rgen Schmidhuber, et~al.
  2001.
\newblock Gradient flow in recurrent nets: the difficulty of learning long-term
  dependencies.

\bibitem[{Hochreiter and Schmidhuber(1997)}]{hochreiter1997long}
Sepp Hochreiter and J{\"u}rgen Schmidhuber. 1997.
\newblock Long short-term memory.
\newblock \emph{Neural computation}, 9(8):1735--1780.

\bibitem[{Huang and Sagae(2010)}]{huang10}
Liang Huang and Kenji Sagae. 2010.
\newblock Dynamic programming for linear-time incremental parsing.
\newblock In \emph{Proceedings of the 48th Annual Meeting of the Association
  for Computational Linguistics ({ACL})}, pages 1077--1086.

\bibitem[{Jawahar et~al.(2018)Jawahar, Muller, Fethi, Martin, Villemonte de~la
  Clergerie, Sagot, and Seddah}]{jawahar18}
Ganesh Jawahar, Benjamin Muller, Amal Fethi, Louis Martin, Eric Villemonte
  de~la Clergerie, Beno{\^\i}t Sagot, and Djam{\'e} Seddah. 2018.
\newblock {ELM}o{L}ex: Connecting {ELM}o and lexicon features for dependency
  parsing.
\newblock In \emph{Proceedings of the {C}o{NLL} 2018 Shared Task: Multilingual
  Parsing from Raw Text to Universal Dependencies}, pages 223--237.

\bibitem[{Johansson and Nugues(2007)}]{johansson07}
Richard Johansson and Pierre Nugues. 2007.
\newblock Incremental dependency parsing using online learning.
\newblock In \emph{Proceedings of the CoNLL Shared Task of EMNLP-CoNLL 2007},
  pages 1134--1138.

\bibitem[{Kiperwasser and Goldberg(2016)}]{kiperwasser16}
Eliyahu Kiperwasser and Yoav Goldberg. 2016.
\newblock Simple and accurate dependency parsing using bidirectional lstm
  feature representations.
\newblock \emph{Transactions of the Association for Computational Linguistics},
  4:313--327.

\bibitem[{Kondratyuk(2019)}]{kondratyuk19}
Daniel Kondratyuk. 2019.
\newblock 75 languages, 1 model: Parsing universal dependencies universally.
\newblock \emph{CoRR}, abs/1904.02099.

\bibitem[{Koo and Collins(2010)}]{koo10acl}
Terry Koo and Michael Collins. 2010.
\newblock Efficient third-order dependency parsers.
\newblock In \emph{Proceedings of the 48th Annual Meeting of the Association
  for Computational Linguistics ({ACL})}, pages 1--11.

\bibitem[{Koo et~al.(2010)Koo, Rush, Collins, Jaakkola, and
  Sontag}]{koo10emnlp}
Terry Koo, Alexander~M. Rush, Michael Collins, Tommi Jaakkola, and David
  Sontag. 2010.
\newblock Dual decomposition for parsing with non-projective head automata.
\newblock In \emph{Proceedings of the 2010 Conference on Empirical Methods in
  Natural Language Processing}, pages 1288--1298.

\bibitem[{Kuhlmann et~al.(2011)Kuhlmann, G\'{o}mez-Rodr\'{i}guez, and
  Satta}]{kuhlmann11}
Marco Kuhlmann, Carlos G\'{o}mez-Rodr\'{i}guez, and Giorgio Satta. 2011.
\newblock Dynamic programming algorithms for transition-based dependency
  parsers.
\newblock In \emph{Proceedings of the 49th Annual Meeting of the Association
  for Computational Linguistics ({ACL})}, pages 673--682.

\bibitem[{de~Lhoneux et~al.(2017{\natexlab{a}})de~Lhoneux, Shao, Basirat,
  Kiperwasser, Stymne, Goldberg, and Nivre}]{delhoneux17conll}
Miryam de~Lhoneux, Yan Shao, Ali Basirat, Eliyahu Kiperwasser, Sara Stymne,
  Yoav Goldberg, and Joakim Nivre. 2017{\natexlab{a}}.
\newblock From raw text to {U}niversal {D}ependencies -- {L}ook, no tags!
\newblock In \emph{Proceedings of the CoNLL 2017 Shared Task: Multilingual
  Parsing from Raw Text to Universal Dependencies}, pages 207--217.

\bibitem[{de~Lhoneux et~al.(2017{\natexlab{b}})de~Lhoneux, Stymne, and
  Nivre}]{delhoneux17iwpt}
Miryam de~Lhoneux, Sara Stymne, and Joakim Nivre. 2017{\natexlab{b}}.
\newblock Arc-hybrid non-projective dependency parsing with a static-dynamic
  oracle.
\newblock In \emph{Proceedings of the 15th International Conference on Parsing
  Technologies}, pages 99--104.

\bibitem[{de~Lhoneux et~al.(2017{\natexlab{c}})de~Lhoneux, Stymne, and
  Nivre}]{delhoneux2017old}
Miryam de~Lhoneux, Sara Stymne, and Joakim Nivre. 2017{\natexlab{c}}.
\newblock Old school vs. new school: Comparing transition-based parsers with
  and without neural network enhancement.
\newblock In \emph{Proceedings of the 15th Treebanks and Linguistic Theories
  Workshop (TLT)}.

\bibitem[{Lim et~al.(2018)Lim, Park, Lee, and Poibeau}]{lim18}
KyungTae Lim, Cheoneum Park, Changki Lee, and Thierry Poibeau. 2018.
\newblock {SE}x {B}i{ST}: A multi-source trainable parser with deep
  contextualized lexical representations.
\newblock In \emph{Proceedings of the {C}o{NLL} 2018 Shared Task: Multilingual
  Parsing from Raw Text to Universal Dependencies}, pages 143--152.

\bibitem[{Liu et~al.(2019)Liu, Gardner, Belinkov, Peters, and Smith}]{liu19}
Nelson~F. Liu, Matt Gardner, Yonatan Belinkov, Matthew~E. Peters, and Noah~A.
  Smith. 2019.
\newblock Linguistic knowledge and transferability of contextual
  representations.
\newblock \emph{CoRR}, abs/1903.08855.

\bibitem[{McDonald et~al.(2005{\natexlab{a}})McDonald, Crammer, and
  Pereira}]{mcdonald05acl}
Ryan McDonald, Koby Crammer, and Fernando Pereira. 2005{\natexlab{a}}.
\newblock Online large-margin training of dependency parsers.
\newblock In \emph{Proceedings of the 43rd Annual Meeting of the Association
  for Computational Linguistics ({ACL})}, pages 91--98.

\bibitem[{McDonald et~al.(2006)McDonald, Lerman, and Pereira}]{mcdonald06conll}
Ryan McDonald, Kevin Lerman, and Fernando Pereira. 2006.
\newblock Multilingual dependency analysis with a two-stage discriminative
  parser.
\newblock In \emph{Proceedings of the 10th Conference on Computational Natural
  Language Learning ({CoNLL})}, pages 216--220.

\bibitem[{McDonald and Nivre(2007)}]{mcdonald07emnlp}
Ryan McDonald and Joakim Nivre. 2007.
\newblock Characterizing the errors of data-driven dependency parsing models.
\newblock In \emph{Proceedings of the 2007 Joint Conference on Empirical
  Methods in Natural Language Processing and Computational Natural Language
  Learning ({EMNLP-CoNLL})}, pages 122--131.

\bibitem[{McDonald and Nivre(2011)}]{mcdonald11cl}
Ryan McDonald and Joakim Nivre. 2011.
\newblock Analyzing and integrating dependency parsers.
\newblock \emph{Computational Linguistics}, pages 197--230.

\bibitem[{McDonald and Pereira(2006)}]{mcdonald06eacl}
Ryan McDonald and Fernando Pereira. 2006.
\newblock Online learning of approximate dependency parsing algorithms.
\newblock In \emph{Proceedings of the 11th Conference of the European Chapter
  of the Association for Computational Linguistics ({EACL})}, pages 81--88.

\bibitem[{McDonald et~al.(2005{\natexlab{b}})McDonald, Pereira, Ribarov, and
  Haji\v{c}}]{mcdonald05emnlp}
Ryan McDonald, Fernando Pereira, Kiril Ribarov, and Jan Haji\v{c}.
  2005{\natexlab{b}}.
\newblock Non-projective dependency parsing using spanning tree algorithms.
\newblock In \emph{Proceedings of the Human Language Technology Conference and
  the Conference on Empirical Methods in Natural Language Processing
  (HLT/EMNLP)}, pages 523--530.

\bibitem[{Mikolov et~al.(2013)Mikolov, Chen, Corrado, and Dean}]{mikolov13}
Tomas Mikolov, Kai Chen, Greg Corrado, and Jeffrey Dean. 2013.
\newblock Efficient estimation of word representations in vector space.
\newblock \emph{arXiv preprint arXiv:1301.3781}.

\bibitem[{Nivre(2003)}]{nivre03iwpt}
Joakim Nivre. 2003.
\newblock An efficient algorithm for projective dependency parsing.
\newblock In \emph{Proceedings of the 8th International Workshop on Parsing
  Technologies ({IWPT})}, pages 149--160.

\bibitem[{Nivre(2008)}]{nivre08cl}
Joakim Nivre. 2008.
\newblock Algorithms for deterministic incremental dependency parsing.
\newblock \emph{Computational Linguistics}, 34:513--553.

\bibitem[{Nivre(2009)}]{nivre09acl}
Joakim Nivre. 2009.
\newblock Non-projective dependency parsing in expected linear time.
\newblock In \emph{Proceedings of the Joint Conference of the 47th Annual
  Meeting of the ACL and the 4th International Joint Conference on Natural
  Language Processing of the AFNLP (ACL-IJCNLP)}, pages 351--359.

\bibitem[{Nivre et~al.(2018)Nivre, Abrams, Agi{\'c}, Ahrenberg, Antonsen,
  Aplonova, Aranzabe, Arutie, Asahara, Ateyah, Attia, Atutxa, Augustinus,
  Badmaeva, Ballesteros, Banerjee, Bank, Barbu~Mititelu, Basmov, Bauer,
  Bellato, Bengoetxea, Berzak, Bhat, Bhat, Biagetti, Bick, Blokland, Bobicev,
  B{\"o}rstell, Bosco, Bouma, Bowman, Boyd, Burchardt, Candito, Caron, Caron,
  Cebiro{\u g}lu~Eryi{\u g}it, Cecchini, Celano, {\v C}{\'e}pl{\"o}, Cetin,
  Chalub, Choi, Cho, Chun, Cinkov{\'a}, Collomb, {\c C}{\"o}ltekin, Connor,
  Courtin, Davidson, de~Marneffe, de~Paiva, Diaz~de Ilarraza, Dickerson, Dirix,
  Dobrovoljc, Dozat, Droganova, Dwivedi, Eli, Elkahky, Ephrem, Erjavec,
  Etienne, Farkas, Fernandez~Alcalde, Foster, Freitas, Gajdo{\v s}ov{\'a},
  Galbraith, Garcia, G{\"a}rdenfors, Garza, Gerdes, Ginter, Goenaga, Gojenola,
  G{\"o}k{\i}rmak, Goldberg, G{\'o}mez~Guinovart, Gonz{\'a}les~Saavedra,
  Grioni, Gr{\=u}z{\={\i}}tis, Guillaume, Guillot-Barbance, Habash, Haji{\v c},
  Haji{\v c}~jr., H{\`a}~M{\~y}, Han, Harris, Haug, Hladk{\'a}, Hlav{\'a}{\v
  c}ov{\'a}, Hociung, Hohle, Hwang, Ion, Irimia, Ishola, Jel{\'{\i}}nek,
  Johannsen, J{\o}rgensen, Ka{\c s}{\i}kara, Kahane, Kanayama, Kanerva, Katz,
  Kayadelen, Kenney, Kettnerov{\'a}, Kirchner, Kopacewicz, Kotsyba, Krek, Kwak,
  Laippala, Lambertino, Lam, Lando, Larasati, Lavrentiev, Lee,
  L{\^e}~H{\`{\^o}}ng, Lenci, Lertpradit, Leung, Li, Li, Li, Lim, Ljube{\v
  s}i{\'c}, Loginova, Lyashevskaya, Lynn, Macketanz, Makazhanov, Mandl,
  Manning, Manurung, M{\u a}r{\u a}nduc, Mare{\v c}ek, Marheinecke,
  Mart{\'{\i}}nez~Alonso, Martins, Ma{\v s}ek, Matsumoto, {McDonald}, Mendon{\c
  c}a, Miekka, Misirpashayeva, Missil{\"a}, Mititelu, Miyao, Montemagni, More,
  Moreno~Romero, Mori, Mori, Mortensen, Moskalevskyi, Muischnek, Murawaki,
  M{\"u}{\"u}risep, Nainwani, Navarro~Hor{\~n}iacek, Nedoluzhko, Ne{\v
  s}pore-B{\=e}rzkalne, Nguy{\~{\^e}}n~Th{\d i}, Nguy{\~{\^e}}n Th{\d i}~Minh,
  Nikolaev, Nitisaroj, Nurmi, Ojala, Ol{\'u}{\`o}kun, Omura, Osenova,
  {\"O}stling, {\O}vrelid, Partanen, Pascual, Passarotti, Patejuk,
  Paulino-Passos, Peng, Perez, Perrier, Petrov, Piitulainen, Pitler, Plank,
  Poibeau, Popel, Pretkalni{\c n}a, Pr{\'e}vost, Prokopidis,
  Przepi{\'o}rkowski, Puolakainen, Pyysalo, R{\"a}{\"a}bis, Rademaker,
  Ramasamy, Rama, Ramisch, Ravishankar, Real, Reddy, Rehm, Rie{\ss}ler,
  Rinaldi, Rituma, Rocha, Romanenko, Rosa, Rovati, Roșca, Rudina, Rueter,
  Sadde, Sagot, Saleh, Samard{\v z}i{\'c}, Samson, Sanguinetti, Saul{\={\i}}te,
  Sawanakunanon, Schneider, Schuster, Seddah, Seeker, Seraji, Shen, Shimada,
  Shohibussirri, Sichinava, Silveira, Simi, Simionescu, Simk{\'o}, {\v
  S}imkov{\'a}, Simov, Smith, Soares-Bastos, Spadine, Stella, Straka,
  Strnadov{\'a}, Suhr, Sulubacak, Sz{\'a}nt{\'o}, Taji, Takahashi, Tanaka,
  Tellier, Trosterud, Trukhina, Tsarfaty, Tyers, Uematsu, Ure{\v s}ov{\'a},
  Uria, Uszkoreit, Vajjala, van Niekerk, van Noord, Varga, Villemonte de~la
  Clergerie, Vincze, Wallin, Wang, Washington, Williams, Wir{\'e}n,
  Woldemariam, Wong, Yan, Yavrumyan, Yu, {\v Z}abokrtsk{\'y}, Zeldes, Zeman,
  Zhang, and Zhu}]{ud23}
Joakim Nivre, Mitchell Abrams, {\v Z}eljko Agi{\'c}, Lars Ahrenberg, Lene
  Antonsen, Katya Aplonova, Maria~Jesus Aranzabe, Gashaw Arutie, Masayuki
  Asahara, Luma Ateyah, Mohammed Attia, Aitziber Atutxa, Liesbeth Augustinus,
  Elena Badmaeva, Miguel Ballesteros, Esha Banerjee, Sebastian Bank, Verginica
  Barbu~Mititelu, Victoria Basmov, John Bauer, Sandra Bellato, Kepa Bengoetxea,
  Yevgeni Berzak, Irshad~Ahmad Bhat, Riyaz~Ahmad Bhat, Erica Biagetti, Eckhard
  Bick, Rogier Blokland, Victoria Bobicev, Carl B{\"o}rstell, Cristina Bosco,
  Gosse Bouma, Sam Bowman, Adriane Boyd, Aljoscha Burchardt, Marie Candito,
  Bernard Caron, Gauthier Caron, G{\"u}l{\c s}en Cebiro{\u g}lu~Eryi{\u g}it,
  Flavio~Massimiliano Cecchini, Giuseppe G.~A. Celano, Slavom{\'{\i}}r {\v
  C}{\'e}pl{\"o}, Savas Cetin, Fabricio Chalub, Jinho Choi, Yongseok Cho,
  Jayeol Chun, Silvie Cinkov{\'a}, Aur{\'e}lie Collomb, {\c C}a{\u g}r{\i} {\c
  C}{\"o}ltekin, Miriam Connor, Marine Courtin, Elizabeth Davidson,
  Marie-Catherine de~Marneffe, Valeria de~Paiva, Arantza Diaz~de Ilarraza,
  Carly Dickerson, Peter Dirix, Kaja Dobrovoljc, Timothy Dozat, Kira Droganova,
  Puneet Dwivedi, Marhaba Eli, Ali Elkahky, Binyam Ephrem, Toma{\v z} Erjavec,
  Aline Etienne, Rich{\'a}rd Farkas, Hector Fernandez~Alcalde, Jennifer Foster,
  Cl{\'a}udia Freitas, Katar{\'{\i}}na Gajdo{\v s}ov{\'a}, Daniel Galbraith,
  Marcos Garcia, Moa G{\"a}rdenfors, Sebastian Garza, Kim Gerdes, Filip Ginter,
  Iakes Goenaga, Koldo Gojenola, Memduh G{\"o}k{\i}rmak, Yoav Goldberg, Xavier
  G{\'o}mez~Guinovart, Berta Gonz{\'a}les~Saavedra, Matias Grioni, Normunds
  Gr{\=u}z{\={\i}}tis, Bruno Guillaume, C{\'e}line Guillot-Barbance, Nizar
  Habash, Jan Haji{\v c}, Jan Haji{\v c}~jr., Linh H{\`a}~M{\~y}, Na-Rae Han,
  Kim Harris, Dag Haug, Barbora Hladk{\'a}, Jaroslava Hlav{\'a}{\v c}ov{\'a},
  Florinel Hociung, Petter Hohle, Jena Hwang, Radu Ion, Elena Irimia, {\d
  O}l{\'a}j{\'{\i}}d{\'e} Ishola, Tom{\'a}{\v s} Jel{\'{\i}}nek, Anders
  Johannsen, Fredrik J{\o}rgensen, H{\"u}ner Ka{\c s}{\i}kara, Sylvain Kahane,
  Hiroshi Kanayama, Jenna Kanerva, Boris Katz, Tolga Kayadelen, Jessica Kenney,
  V{\'a}clava Kettnerov{\'a}, Jesse Kirchner, Kamil Kopacewicz, Natalia
  Kotsyba, Simon Krek, Sookyoung Kwak, Veronika Laippala, Lorenzo Lambertino,
  Lucia Lam, Tatiana Lando, Septina~Dian Larasati, Alexei Lavrentiev, John Lee,
  Phuong L{\^e}~H{\`{\^o}}ng, Alessandro Lenci, Saran Lertpradit, Herman Leung,
  Cheuk~Ying Li, Josie Li, Keying Li, {KyungTae} Lim, Nikola Ljube{\v s}i{\'c},
  Olga Loginova, Olga Lyashevskaya, Teresa Lynn, Vivien Macketanz, Aibek
  Makazhanov, Michael Mandl, Christopher Manning, Ruli Manurung, C{\u a}t{\u
  a}lina M{\u a}r{\u a}nduc, David Mare{\v c}ek, Katrin Marheinecke, H{\'e}ctor
  Mart{\'{\i}}nez~Alonso, Andr{\'e} Martins, Jan Ma{\v s}ek, Yuji Matsumoto,
  Ryan {McDonald}, Gustavo Mendon{\c c}a, Niko Miekka, Margarita
  Misirpashayeva, Anna Missil{\"a}, C{\u a}t{\u a}lin Mititelu, Yusuke Miyao,
  Simonetta Montemagni, Amir More, Laura Moreno~Romero, Keiko~Sophie Mori,
  Shinsuke Mori, Bjartur Mortensen, Bohdan Moskalevskyi, Kadri Muischnek, Yugo
  Murawaki, Kaili M{\"u}{\"u}risep, Pinkey Nainwani, Juan~Ignacio
  Navarro~Hor{\~n}iacek, Anna Nedoluzhko, Gunta Ne{\v s}pore-B{\=e}rzkalne,
  Luong Nguy{\~{\^e}}n~Th{\d i}, Huy{\`{\^e}}n Nguy{\~{\^e}}n Th{\d i}~Minh,
  Vitaly Nikolaev, Rattima Nitisaroj, Hanna Nurmi, Stina Ojala, Ad{\'e}day{\d
  o} Ol{\'u}{\`o}kun, Mai Omura, Petya Osenova, Robert {\"O}stling, Lilja
  {\O}vrelid, Niko Partanen, Elena Pascual, Marco Passarotti, Agnieszka
  Patejuk, Guilherme Paulino-Passos, Siyao Peng, Cenel-Augusto Perez, Guy
  Perrier, Slav Petrov, Jussi Piitulainen, Emily Pitler, Barbara Plank, Thierry
  Poibeau, Martin Popel, Lauma Pretkalni{\c n}a, Sophie Pr{\'e}vost, Prokopis
  Prokopidis, Adam Przepi{\'o}rkowski, Tiina Puolakainen, Sampo Pyysalo,
  Andriela R{\"a}{\"a}bis, Alexandre Rademaker, Loganathan Ramasamy, Taraka
  Rama, Carlos Ramisch, Vinit Ravishankar, Livy Real, Siva Reddy, Georg Rehm,
  Michael Rie{\ss}ler, Larissa Rinaldi, Laura Rituma, Luisa Rocha, Mykhailo
  Romanenko, Rudolf Rosa, Davide Rovati, Valentin Roșca, Olga Rudina, Jack
  Rueter, Shoval Sadde, Beno{\^{\i}}t Sagot, Shadi Saleh, Tanja Samard{\v
  z}i{\'c}, Stephanie Samson, Manuela Sanguinetti, Baiba Saul{\={\i}}te, Yanin
  Sawanakunanon, Nathan Schneider, Sebastian Schuster, Djam{\'e} Seddah,
  Wolfgang Seeker, Mojgan Seraji, Mo~Shen, Atsuko Shimada, Muh Shohibussirri,
  Dmitry Sichinava, Natalia Silveira, Maria Simi, Radu Simionescu, Katalin
  Simk{\'o}, M{\'a}ria {\v S}imkov{\'a}, Kiril Simov, Aaron Smith, Isabela
  Soares-Bastos, Carolyn Spadine, Antonio Stella, Milan Straka, Jana
  Strnadov{\'a}, Alane Suhr, Umut Sulubacak, Zsolt Sz{\'a}nt{\'o}, Dima Taji,
  Yuta Takahashi, Takaaki Tanaka, Isabelle Tellier, Trond Trosterud, Anna
  Trukhina, Reut Tsarfaty, Francis Tyers, Sumire Uematsu, Zde{\v n}ka Ure{\v
  s}ov{\'a}, Larraitz Uria, Hans Uszkoreit, Sowmya Vajjala, Daniel van Niekerk,
  Gertjan van Noord, Viktor Varga, Eric Villemonte de~la Clergerie, Veronika
  Vincze, Lars Wallin, Jing~Xian Wang, Jonathan~North Washington, Seyi
  Williams, Mats Wir{\'e}n, Tsegay Woldemariam, Tak-sum Wong, Chunxiao Yan,
  Marat~M. Yavrumyan, Zhuoran Yu, Zden{\v e}k {\v Z}abokrtsk{\'y}, Amir Zeldes,
  Daniel Zeman, Manying Zhang, and Hanzhi Zhu. 2018.
\newblock \href {http://hdl.handle.net/11234/1-2895} {Universal dependencies
  2.3}.
\newblock {LINDAT}/{CLARIN} digital library at the Institute of Formal and
  Applied Linguistics ({{\'U}FAL}), Faculty of Mathematics and Physics, Charles
  University.

\bibitem[{Nivre et~al.(2007)Nivre, Hall, K\"ubler, McDonald, Nilsson, Riedel,
  and Yuret}]{nivre07conll}
Joakim Nivre, Johan Hall, Sandra K\"ubler, Ryan McDonald, Jens Nilsson,
  Sebastian Riedel, and Deniz Yuret. 2007.
\newblock The {CoNLL} 2007 shared task on dependency parsing.
\newblock In \emph{Proceedings of the CoNLL Shared Task of EMNLP-CoNLL 2007},
  pages 915--932.

\bibitem[{Nivre et~al.(2006)Nivre, Hall, Nilsson, Eryi{\u g}it, and
  Marinov}]{nivre06conll}
Joakim Nivre, Johan Hall, Jens Nilsson, G{\"u}lsen Eryi{\u g}it, and Svetoslav
  Marinov. 2006.
\newblock Labeled pseudo-projective dependency parsing with support vector
  machines.
\newblock In \emph{Proceedings of the 10th Conference on Computational Natural
  Language Learning ({CoNLL})}, pages 221--225.

\bibitem[{Nivre and McDonald(2008)}]{nivre08acl}
Joakim Nivre and Ryan McDonald. 2008.
\newblock Integrating graph-based and transition-based dependency parsers.
\newblock In \emph{Proceedings of the 46th Annual Meeting of the Association
  for Computational Linguistics ({ACL})}, pages 950--958.

\bibitem[{Nivre and Nilsson(2005)}]{nivre05acl}
Joakim Nivre and Jens Nilsson. 2005.
\newblock Pseudo-projective dependency parsing.
\newblock In \emph{Proceedings of the 43rd Annual Meeting of the Association
  for Computational Linguistics ({ACL})}, pages 99--106.

\bibitem[{Pennington et~al.(2014)Pennington, Socher, and
  Manning}]{pennington14}
Jeffrey Pennington, Richard Socher, and Christopher Manning. 2014.
\newblock {G}love: Global vectors for word representation.
\newblock In \emph{Proceedings of the Conference on Empirical Methods in
  Natural Language Processing ({EMNLP})}, pages 1532--1543.

\bibitem[{Peters et~al.(2018)Peters, Neumann, Iyyer, Gardner, clark, Lee, and
  Zettlemoyer}]{peters18}
Matthew~E. Peters, Mark Neumann, Mohit Iyyer, Matt Gardner, Christopher clark,
  Kenton Lee, and Luke Zettlemoyer. 2018.
\newblock Deep contextualized word representations.
\newblock In \emph{Proceedings of the 2018 Conference of the North {A}merican
  Chapter of the Association for Computational Linguistics: Human Language
  Technologies, Volume 1 (Long Papers)}, pages 2227--2237.

\bibitem[{Qi et~al.(2018)Qi, Dozat, Zhang, and Manning}]{qi2018universal}
Peng Qi, Timothy Dozat, Yuhao Zhang, and Christopher~D Manning. 2018.
\newblock Universal dependency parsing from scratch.
\newblock In \emph{Proceedings of the 2018 CoNLL Shared Task: Multilingual
  Parsing from Raw Text to Universal Dependencies}, page 160.

\bibitem[{Sagae and Lavie(2006)}]{sagae06naacl}
Kenji Sagae and Alon Lavie. 2006.
\newblock Parser combination by reparsing.
\newblock In \emph{Proceedings of the Human Language Technology Conference of
  the NAACL, Companion Volume: Short Papers}, pages 129--132.

\bibitem[{Schuster et~al.(2019)Schuster, Ram, Barzilay, and
  Globerson}]{schuster19crosslingual}
Tal Schuster, Ori Ram, Regina Barzilay, and Amir Globerson. 2019.
\newblock \href {https://doi.org/10.18653/v1/N19-1162} {Cross-lingual alignment
  of contextual word embeddings, with applications to zero-shot dependency
  parsing}.
\newblock In \emph{Proceedings of the 2019 Conference of the North {A}merican
  Chapter of the Association for Computational Linguistics: Human Language
  Technologies, Volume 1 (Long and Short Papers)}, pages 1599--1613,
  Minneapolis, Minnesota. Association for Computational Linguistics.

\bibitem[{Smith et~al.(2018{\natexlab{a}})Smith, Bohnet, de~Lhoneux, Nivre,
  Shao, and Stymne}]{smith18conll}
Aaron Smith, Bernd Bohnet, Miryam de~Lhoneux, Joakim Nivre, Yan Shao, and Sara
  Stymne. 2018{\natexlab{a}}.
\newblock 82 treebanks, 34 models: Universal dependency parsing with
  multi-treebank models.
\newblock In \emph{Proceedings of the 2018 CoNLL Shared Task: Multilingual
  Parsing from Raw Text to Universal Dependencies}.

\bibitem[{Smith et~al.(2018{\natexlab{b}})Smith, de~Lhoneux, Stymne, and
  Nivre}]{smith18emnlp}
Aaron Smith, Miryam de~Lhoneux, Sara Stymne, and Joakim Nivre.
  2018{\natexlab{b}}.
\newblock An investigation of the interactions between pre-trained word
  embeddings, character models and pos tags in dependency parsing.
\newblock In \emph{Proceedings of the 2018 Conference on Empirical Methods in
  Natural Language Processing}.

\bibitem[{Tenney et~al.(2019)Tenney, Xia, Chen, Wang, Poliak, McCoy, Kim,
  Van~Durme, Bowman, Das, and Pavlick}]{tenney19}
Ian Tenney, Patrick Xia, Berlin Chen, Alex Wang, Adam Poliak, R.~Thomas McCoy,
  Najoung Kim, Benjamin Van~Durme, Samuel~R. Bowman, Dipanjan Das, and Ellie
  Pavlick. 2019.
\newblock What do you learn from context? probing for sentence structure in
  contextualized word representations.
\newblock In \emph{Proceedings of the 5th International Conference on Learning
  Representations}.

\bibitem[{Titov and Henderson(2007)}]{titov07iwpt}
Ivan Titov and James Henderson. 2007.
\newblock A latent variable model for generative dependency parsing.
\newblock In \emph{Proceedings of the 10th International Conference on Parsing
  Technologies ({IWPT})}, pages 144--155.

\bibitem[{Tran et~al.(2018)Tran, Bisazza, and Monz}]{tran2018importance}
Ke~Tran, Arianna Bisazza, and Christof Monz. 2018.
\newblock The importance of being recurrent for modeling hierarchical
  structure.
\newblock In \emph{Proceedings of the 2018 Conference on Empirical Methods in
  Natural Language Processing}, pages 4731--4736.

\bibitem[{Vaswani et~al.(2017)Vaswani, Shazeer, Parmar, Uszkoreit, Jones,
  Gomez, Kaiser, and Polosukhin}]{vaswani2017attention}
Ashish Vaswani, Noam Shazeer, Niki Parmar, Jakob Uszkoreit, Llion Jones,
  Aidan~N Gomez, {\L}ukasz Kaiser, and Illia Polosukhin. 2017.
\newblock Attention is all you need.
\newblock In \emph{Advances in Neural Information Processing Systems}, pages
  5998--6008.

\bibitem[{Veenstra and Daelemans(2000)}]{veenstra00}
Jorn Veenstra and Walter Daelemans. 2000.
\newblock A memory-based alternative for connectionist shift-reduce parsing.
\newblock Technical Report ILK-0012, Tilburg University.

\bibitem[{Wang et~al.(2019)Wang, Singh, Michael, Hill, Levy, and
  Bowman}]{wang2019glue}
Alex Wang, Amanpreet Singh, Julian Michael, Felix Hill, Omer Levy, and
  Samuel~R. Bowman. 2019.
\newblock {GLUE}: A multi-task benchmark and analysis platform for natural
  language understanding.
\newblock In \emph{Proceedings of the 7th International Conference on Learning
  Representations}.

\bibitem[{Weiss et~al.(2015)Weiss, Alberti, Collins, and Petrov}]{weiss15}
David Weiss, Chris Alberti, Michael Collins, and Slav Petrov. 2015.
\newblock Structured training for neural network transition-based parsing.
\newblock In \emph{Proceedings of the 53rd Annual Meeting of the Association
  for Computational Linguistics ({ACL})}, pages 323--333.

\bibitem[{Wu et~al.(2016)Wu, Schuster, Chen, Le, Norouzi, Macherey, Krikun,
  Cao, Gao, Macherey et~al.}]{wu2016google}
Yonghui Wu, Mike Schuster, Zhifeng Chen, Quoc~V Le, Mohammad Norouzi, Wolfgang
  Macherey, Maxim Krikun, Yuan Cao, Qin Gao, Klaus Macherey, et~al. 2016.
\newblock Google's neural machine translation system: Bridging the gap between
  human and machine translation.
\newblock \emph{arXiv preprint arXiv:1609.08144}.

\bibitem[{Yamada and Matsumoto(2003)}]{yamada03}
Hiroyasu Yamada and Yuji Matsumoto. 2003.
\newblock Statistical dependency analysis with support vector machines.
\newblock In \emph{Proceedings of the 8th International Workshop on Parsing
  Technologies ({IWPT})}, pages 195--206.

\bibitem[{Zeman et~al.(2018)Zeman, Haji{\v{c}}, Popel, Potthtyersast, Straka,
  Ginter, Nivre, and Petrov}]{zeman18conll}
Daniel Zeman, Jan Haji{\v{c}}, Martin Popel, Martin Potthtyersast, Milan
  Straka, Filip Ginter, Joakim Nivre, and Slav Petrov. 2018.
\newblock {CoNLL 2018 Shared Task: Multilingual Parsing from Raw Text to
  Universal Dependencies}.
\newblock In \emph{Proceedings of the CoNLL 2018 Shared Task: Multilingual
  Parsing from Raw Text to Universal Dependencies}.

\bibitem[{Zhang and McDonald(2012)}]{zhang12mcdonald}
Hao Zhang and Ryan McDonald. 2012.
\newblock Generalized higher-order dependency parsing with cube pruning.
\newblock In \emph{Proceedings of the 2012 Joint Conference on Empirical
  Methods in Natural Language Processing and Computational Natural Language
  Learning ({EMNLP-CoNLL})}, pages 320--331.

\bibitem[{Zhang and Clark(2008)}]{zhang08emnlp}
Yue Zhang and Stephen Clark. 2008.
\newblock A tale of two parsers: {I}nvestigating and combining graph-based and
  transition-based dependency parsing.
\newblock In \emph{Proceedings of the Conference on Empirical Methods in
  Natural Language Processing ({EMNLP})}, pages 562--571.

\bibitem[{Zhang and Nivre(2011)}]{zhang11}
Yue Zhang and Joakim Nivre. 2011.
\newblock Transition-based parsing with rich non-local features.
\newblock In \emph{Proceedings of the 49th Annual Meeting of the Association
  for Computational Linguistics ({ACL})}, pages 188--193.

\bibitem[{Zhang and Nivre(2012)}]{zhang12}
Yue Zhang and Joakim Nivre. 2012.
\newblock Analyzing the effect of global learning and beam-search on
  transition-based dependency parsing.
\newblock In \emph{Proceedings of COLING 2012: Posters}, pages 1391--1400.

\end{thebibliography}
\clearpage
%
%
%
%
%
%
\appendix

\begin{center}%
{\LARGE\bfseries Supplementary Material}
\end{center}

\section{Hyperparameters}
\label{sec:appa}
\begin{table}[ht!]
\begin{tabular}{@{}rl@{}}
\toprule
\multicolumn{1}{r}{\textbf{Component}} & \multicolumn{1}{l}{\textbf{Specification}} \\ \midrule
$\mathbf{w}_k$ dim.                              & $500$ ($100 * 2$ (char) $+$ $300$ (word)) \\
$\mathbf{ELMo}_k$ dim.                               & $1024$                                \\
$\mathbf{BERT}_k$ dim.                              & $768$                                 \\
BiLSTM output dim.                     & $125$                                 \\
MLP output dim.                        & $100$                                 \\
No. BiLSTM layers                      & $2$                                   \\
BiLSTM dropout rate                   & $0.33$                                \\
No. training epochs                    & $30$                                  \\
Model selection                        & best dev. performance               \\ \bottomrule
\end{tabular}
\end{table}

\section{Error Analysis}
\label{sec:appb}

\subsection{Dependency Length}

\begin{table}[ht!]
\begin{tabular}{@{}rcccccccccccc@{}}
\toprule
\textbf{}         & \multicolumn{2}{c}{\textsc{gr}} & \multicolumn{2}{c}{\textsc{gr+B}} & \multicolumn{2}{c}{\textsc{gr+E}} & \multicolumn{2}{c}{\textsc{tr}} & \multicolumn{2}{c}{\textsc{tr+B}} & \multicolumn{2}{c}{\textsc{tr+E}} \\ \midrule
                  & pr         & rc        & pr          & rc         & pr          & rc         & pr         & rc        & pr          & rc         & pr          & rc         \\ \midrule
root              & 0.91       & 0.91      & 0.92        & 0.92       & 0.92        & 0.92       & 0.88       & 0.88      & 0.91        & 0.91       & 0.90        & 0.90       \\
1                 & 0.89       & 0.90      & 0.90        & 0.91       & 0.90        & 0.91       & 0.88       & 0.90      & 0.90        & 0.91       & 0.90        & 0.92       \\
2                 & 0.83       & 0.84      & 0.85        & 0.86       & 0.86        & 0.86       & 0.82       & 0.83      & 0.86        & 0.86       & 0.85        & 0.86       \\
3                 & 0.76       & 0.77      & 0.80        & 0.80       & 0.80        & 0.80       & 0.76       & 0.75      & 0.81        & 0.80       & 0.80        & 0.80       \\
4                 & 0.74       & 0.72      & 0.78        & 0.77       & 0.77        & 0.76       & 0.73       & 0.70      & 0.78        & 0.76       & 0.78        & 0.76       \\
5                 & 0.70       & 0.67      & 0.74        & 0.72       & 0.73        & 0.71       & 0.67       & 0.64      & 0.74        & 0.71       & 0.73        & 0.71       \\
6                 & 0.68       & 0.66      & 0.73        & 0.72       & 0.72        & 0.70       & 0.65       & 0.63      & 0.73        & 0.72       & 0.70        & 0.69       \\
7                 & 0.66       & 0.64      & 0.72        & 0.70       & 0.70        & 0.69       & 0.64       & 0.61      & 0.72        & 0.71       & 0.68        & 0.67       \\
8                 & 0.65       & 0.61      & 0.71        & 0.67       & 0.70        & 0.66       & 0.63       & 0.58      & 0.71        & 0.67       & 0.68        & 0.64       \\
9                 & 0.66       & 0.64      & 0.71        & 0.72       & 0.69        & 0.70       & 0.64       & 0.63      & 0.71        & 0.70       & 0.68        & 0.67       \\
\textgreater{}=10 & 0.69       & 0.66      & 0.73        & 0.71       & 0.71        & 0.69       & 0.65       & 0.63      & 0.72        & 0.71       & 0.70        & 0.68       \\ \bottomrule
\end{tabular}
\end{table}
\clearpage

\subsection{Distance to Root}

\begin{table}[ht!]
\begin{tabular}{@{}rcccccccccccc@{}}
\toprule
\textbf{}         & \multicolumn{2}{c}{\textsc{gr}} & \multicolumn{2}{c}{\textsc{gr+B}} & \multicolumn{2}{c}{\textsc{gr+E}} & \multicolumn{2}{c}{\textsc{tr}} & \multicolumn{2}{c}{\textsc{tr+B}} & \multicolumn{2}{c}{\textsc{tr+E}} \\ \midrule
                  & pr         & rc        & pr          & rc         & pr          & rc         & pr         & rc        & pr          & rc         & pr          & rc         \\ \midrule
0                 & 0.91       & 0.91      & 0.92        & 0.92       & 0.92        & 0.92       & 0.88       & 0.88      & 0.91        & 0.91       & 0.90        & 0.90       \\
1                 & 0.83       & 0.81      & 0.85        & 0.84       & 0.85        & 0.84       & 0.81       & 0.80      & 0.85        & 0.84       & 0.84        & 0.83       \\
2                 & 0.81       & 0.82      & 0.84        & 0.85       & 0.84        & 0.84       & 0.81       & 0.81      & 0.84        & 0.85       & 0.84        & 0.84       \\
3                 & 0.81       & 0.82      & 0.85        & 0.85       & 0.84        & 0.84       & 0.81       & 0.81      & 0.85        & 0.85       & 0.84        & 0.84       \\
4                 & 0.82       & 0.82      & 0.85        & 0.85       & 0.84        & 0.84       & 0.81       & 0.82      & 0.85        & 0.85       & 0.85        & 0.85       \\
5                 & 0.82       & 0.83      & 0.85        & 0.86       & 0.84        & 0.85       & 0.81       & 0.83      & 0.85        & 0.86       & 0.84        & 0.85       \\
6                 & 0.82       & 0.83      & 0.84        & 0.86       & 0.84        & 0.85       & 0.82       & 0.83      & 0.85        & 0.86       & 0.84        & 0.85       \\
7                 & 0.82       & 0.83      & 0.85        & 0.86       & 0.85        & 0.85       & 0.81       & 0.83      & 0.85        & 0.86       & 0.85        & 0.85       \\
8                 & 0.82       & 0.84      & 0.83        & 0.85       & 0.84        & 0.84       & 0.81       & 0.83      & 0.84        & 0.85       & 0.85        & 0.85       \\
9                 & 0.83       & 0.81      & 0.83        & 0.82       & 0.84        & 0.83       & 0.79       & 0.81      & 0.84        & 0.83       & 0.83        & 0.82       \\
\textgreater{}=10 & 0.81       & 0.84      & 0.84        & 0.85       & 0.85        & 0.87       & 0.80       & 0.84      & 0.83        & 0.85       & 0.83        & 0.87       \\ \bottomrule
\end{tabular}
\end{table}

\subsection{Projectivity}

\begin{table}[ht!]
\begin{tabular}{@{}rcccccccccccc@{}}
\toprule
\textbf{} & \multicolumn{2}{c}{\textsc{gr}} & \multicolumn{2}{c}{\textsc{gr+B}} & \multicolumn{2}{c}{\textsc{gr+E}} & \multicolumn{2}{c}{\textsc{tr}} & \multicolumn{2}{c}{\textsc{tr+B}} & \multicolumn{2}{c}{\textsc{tr+E}} \\ \midrule
          & pr         & rc        & pr          & rc         & pr          & rc         & pr         & rc        & pr          & rc         & pr          & rc         \\ \midrule
Non-proj  & 0.35       & 0.43      & 0.36        & 0.45       & 0.36        & 0.45       & 0.48       & 0.33      & 0.49        & 0.37       & 0.52        & 0.40       \\
Proj      & 0.83       & 0.83      & 0.86        & 0.85       & 0.85        & 0.85       & 0.81       & 0.82      & 0.85        & 0.85       & 0.85        & 0.85       \\ \bottomrule
\end{tabular}
\end{table}
\subsection{Part of Speech}

\begin{table}[ht!]
\begin{tabular}{@{}rcccccc@{}}
\toprule
\textbf{} & \textsc{gr}   & \textsc{gr+B} & \textsc{gr+E} & \textsc{tr}   & \textsc{tr+B} & \textsc{tr+E} \\ \midrule
\texttt{ADJ}       & 0.82 & 0.85 & 0.85 & 0.82 & 0.85 & 0.85 \\
\texttt{ADP}       & 0.95 & 0.96 & 0.96 & 0.95 & 0.96 & 0.96 \\
\texttt{ADV}       & 0.78 & 0.80 & 0.80 & 0.77 & 0.80 & 0.81 \\
\texttt{AUX}       & 0.93 & 0.93 & 0.94 & 0.92 & 0.93 & 0.94 \\
\texttt{CCONJ}     & 0.85 & 0.88 & 0.87 & 0.84 & 0.87 & 0.86 \\
\texttt{DET}       & 0.94 & 0.95 & 0.95 & 0.94 & 0.95 & 0.95 \\
\texttt{INTJ}      & 0.50 & 0.44 & 0.48 & 0.44 & 0.45 & 0.48 \\
\texttt{NOUN}      & 0.77 & 0.81 & 0.80 & 0.76 & 0.80 & 0.80 \\
\texttt{NUM}       & 0.83 & 0.84 & 0.84 & 0.82 & 0.84 & 0.84 \\
\texttt{PART}      & 0.86 & 0.89 & 0.87 & 0.84 & 0.90 & 0.88 \\
\texttt{PRON}      & 0.83 & 0.85 & 0.86 & 0.82 & 0.85 & 0.86 \\
\texttt{PROPN}     & 0.79 & 0.84 & 0.82 & 0.78 & 0.84 & 0.83 \\
\texttt{PUNCT}     & 0.82 & 0.85 & 0.84 & 0.81 & 0.86 & 0.84 \\
\texttt{SCONJ}     & 0.90 & 0.93 & 0.93 & 0.89 & 0.93 & 0.93 \\
\texttt{SYM}       & 0.85 & 0.84 & 0.84 & 0.82 & 0.85 & 0.86 \\
\texttt{VERB}      & 0.76 & 0.80 & 0.79 & 0.74 & 0.79 & 0.78 \\
\texttt{X}         & 0.64 & 0.67 & 0.65 & 0.64 & 0.68 & 0.67 \\ \bottomrule
\end{tabular}
\end{table}

\clearpage

\subsection{Dependency Relation}

\begin{table}[ht!]
\begin{tabular}{@{}rcccccccccccc@{}}
\toprule
\textbf{}  & \multicolumn{2}{c}{\textsc{gr}} & \multicolumn{2}{c}{\textsc{gr+B}} & \multicolumn{2}{c}{\textsc{gr+E}} & \multicolumn{2}{c}{\textsc{tr}} & \multicolumn{2}{c}{\textsc{tr+B}} & \multicolumn{2}{c}{\textsc{tr+E}} \\ \midrule
           & pr         & rc        & pr          & rc         & pr          & rc         & pr         & rc        & pr          & rc         & pr          & rc         \\ \midrule
\texttt{acl}        & 0.69       & 0.69      & 0.75        & 0.76       & 0.73        & 0.73       & 0.66       & 0.66      & 0.75        & 0.75       & 0.73        & 0.73       \\
\texttt{advcl}      & 0.68       & 0.69      & 0.68        & 0.71       & 0.70        & 0.72       & 0.66       & 0.67      & 0.69        & 0.71       & 0.70        & 0.71       \\
\texttt{advmod}     & 0.77       & 0.77      & 0.79        & 0.78       & 0.80        & 0.79       & 0.76       & 0.76      & 0.79        & 0.80       & 0.80        & 0.80       \\
\texttt{amod}       & 0.86       & 0.87      & 0.88        & 0.89       & 0.89        & 0.90       & 0.85       & 0.87      & 0.88        & 0.89       & 0.88        & 0.90       \\
\texttt{appos}      & 0.55       & 0.54      & 0.61        & 0.63       & 0.59        & 0.60       & 0.54       & 0.57      & 0.62        & 0.64       & 0.59        & 0.61       \\
\texttt{aux}        & 0.95       & 0.95      & 0.95        & 0.95       & 0.96        & 0.96       & 0.95       & 0.95      & 0.95        & 0.95       & 0.96        & 0.96       \\
\texttt{case}       & 0.95       & 0.96      & 0.96        & 0.97       & 0.96        & 0.96       & 0.95       & 0.95      & 0.96        & 0.97       & 0.96        & 0.96       \\
\texttt{cc}         & 0.84       & 0.85      & 0.87        & 0.88       & 0.87        & 0.87       & 0.83       & 0.83      & 0.87        & 0.88       & 0.86        & 0.86       \\
\texttt{ccomp}      & 0.67       & 0.64      & 0.71        & 0.67       & 0.71        & 0.66       & 0.65       & 0.60      & 0.72        & 0.68       & 0.71        & 0.68       \\
\texttt{clf}        & 0.87       & 0.90      & 0.94        & 0.95       & 0.92        & 0.92       & 0.89       & 0.87      & 0.95        & 0.94       & 0.92        & 0.90       \\
\texttt{compound}   & 0.85       & 0.84      & 0.86        & 0.87       & 0.87        & 0.86       & 0.84       & 0.84      & 0.87        & 0.87       & 0.86        & 0.87       \\
\texttt{conj}       & 0.68       & 0.69      & 0.76        & 0.75       & 0.72        & 0.72       & 0.64       & 0.65      & 0.74        & 0.74       & 0.71        & 0.70       \\
\texttt{cop}        & 0.84       & 0.85      & 0.86        & 0.86       & 0.87        & 0.87       & 0.83       & 0.84      & 0.85        & 0.86       & 0.86        & 0.87       \\
\texttt{csubj}      & 0.59       & 0.51      & 0.62        & 0.54       & 0.63        & 0.56       & 0.55       & 0.49      & 0.64        & 0.53       & 0.63        & 0.55       \\
\texttt{dep}        & 0.68       & 0.56      & 0.74        & 0.60       & 0.72        & 0.58       & 0.65       & 0.56      & 0.73        & 0.60       & 0.69        & 0.57       \\
\texttt{det}        & 0.93       & 0.94      & 0.94        & 0.95       & 0.94        & 0.95       & 0.92       & 0.93      & 0.95        & 0.95       & 0.94        & 0.95       \\
\texttt{discourse}  & 0.74       & 0.61      & 0.78        & 0.62       & 0.75        & 0.65       & 0.72       & 0.60      & 0.79        & 0.65       & 0.76        & 0.56       \\
\texttt{dislocated} & 0.40       & 0.32      & 0.43        & 0.32       & 0.44        & 0.39       & 0.28       & 0.26      & 0.34        & 0.36       & 0.35        & 0.44       \\
\texttt{expl}       & 0.75       & 0.81      & 0.79        & 0.82       & 0.79        & 0.81       & 0.75       & 0.79      & 0.79        & 0.83       & 0.81        & 0.82       \\
\texttt{fixed}      & 0.78       & 0.72      & 0.81        & 0.76       & 0.82        & 0.77       & 0.78       & 0.73      & 0.82        & 0.76       & 0.82        & 0.78       \\
\texttt{flat}       & 0.79       & 0.78      & 0.85        & 0.83       & 0.83        & 0.81       & 0.79       & 0.78      & 0.83        & 0.82       & 0.82        & 0.82       \\
\texttt{goeswith}   & 0.71       & 0.44      & 0.92        & 0.31       & 0.68        & 0.44       & 0.61       & 0.28      & 0.84        & 0.41       & 0.75        & 0.46       \\
\texttt{iobj}       & 0.83       & 0.81      & 0.82        & 0.80       & 0.86        & 0.82       & 0.83       & 0.78      & 0.85        & 0.79       & 0.85        & 0.83       \\
\texttt{list}       & 0.31       & 0.28      & 0.32        & 0.32       & 0.38        & 0.33       & 0.28       & 0.23      & 0.34        & 0.33       & 0.41        & 0.36       \\
\texttt{mark}       & 0.89       & 0.87      & 0.91        & 0.91       & 0.91        & 0.91       & 0.88       & 0.87      & 0.91        & 0.91       & 0.91        & 0.90       \\
\texttt{nmod}       & 0.78       & 0.78      & 0.81        & 0.81       & 0.80        & 0.81       & 0.77       & 0.77      & 0.81        & 0.81       & 0.80        & 0.81       \\
\texttt{nsubj}      & 0.79       & 0.80      & 0.82        & 0.83       & 0.82        & 0.83       & 0.77       & 0.78      & 0.82        & 0.83       & 0.81        & 0.82       \\
\texttt{nummod}     & 0.87       & 0.90      & 0.88        & 0.91       & 0.88        & 0.91       & 0.87       & 0.89      & 0.88        & 0.92       & 0.88        & 0.92       \\
\texttt{obj}        & 0.79       & 0.80      & 0.82        & 0.83       & 0.83        & 0.84       & 0.77       & 0.78      & 0.83        & 0.83       & 0.84        & 0.83       \\
\texttt{obl}        & 0.74       & 0.75      & 0.77        & 0.77       & 0.77        & 0.78       & 0.73       & 0.73      & 0.77        & 0.77       & 0.77        & 0.78       \\
\texttt{orphan}     & 0.29       & 0.11      & 0.31        & 0.12       & 0.35        & 0.11       & 0.21       & 0.10      & 0.35        & 0.13       & 0.38        & 0.15       \\
\texttt{parataxis}  & 0.77       & 0.70      & 0.80        & 0.72       & 0.77        & 0.72       & 0.75       & 0.70      & 0.80        & 0.74       & 0.79        & 0.73       \\
\texttt{punct}      & 0.82       & 0.82      & 0.85        & 0.85       & 0.84        & 0.84       & 0.81       & 0.81      & 0.86        & 0.86       & 0.84        & 0.84       \\
\texttt{root}       & 0.90       & 0.90      & 0.92        & 0.92       & 0.92        & 0.92       & 0.90       & 0.88      & 0.92        & 0.91       & 0.92        & 0.90       \\
\texttt{vocative}   & 0.43       & 0.28      & 0.43        & 0.32       & 0.65        & 0.43       & 0.53       & 0.35      & 0.58        & 0.38       & 0.52        & 0.38       \\
\texttt{xcomp}      & 0.68       & 0.62      & 0.71        & 0.66       & 0.70        & 0.68       & 0.65       & 0.62      & 0.74        & 0.69       & 0.73        & 0.70       \\ \bottomrule
\end{tabular}
\end{table}
\clearpage

\subsection{Sentence Length}

\begin{table}[ht!]
\begin{tabular}{@{}rcccccc@{}}
\toprule
\textbf{} & \textsc{gr}    & \textsc{gr+B}  & \textsc{gr+E}  & \textsc{tr}    & \textsc{tr+B}  & \textsc{tr+E}  \\ \midrule
1-10      & 83.50 & 83.40 & 84.90 & 81.80 & 83.40 & 84.80 \\
11-20     & 83.50 & 85.80 & 86.20 & 82.10 & 85.70 & 86.00 \\
21-30     & 83.00 & 86.20 & 85.50 & 82.00 & 86.10 & 85.10 \\
31-40     & 81.90 & 85.20 & 84.30 & 81.00 & 84.90 & 84.20 \\
41-50     & 81.70 & 84.40 & 83.80 & 80.70 & 84.10 & 83.60 \\
50+       & 78.70 & 81.60 & 81.50 & 78.60 & 82.20 & 81.70 \\ \bottomrule
\end{tabular}
\end{table}
\clearpage
\onecolumn

\section{Per-Language Error Analysis}
\label{sec:appc}

\begin{figure}[ht!]
\centering
\includegraphics[width=\columnwidth]{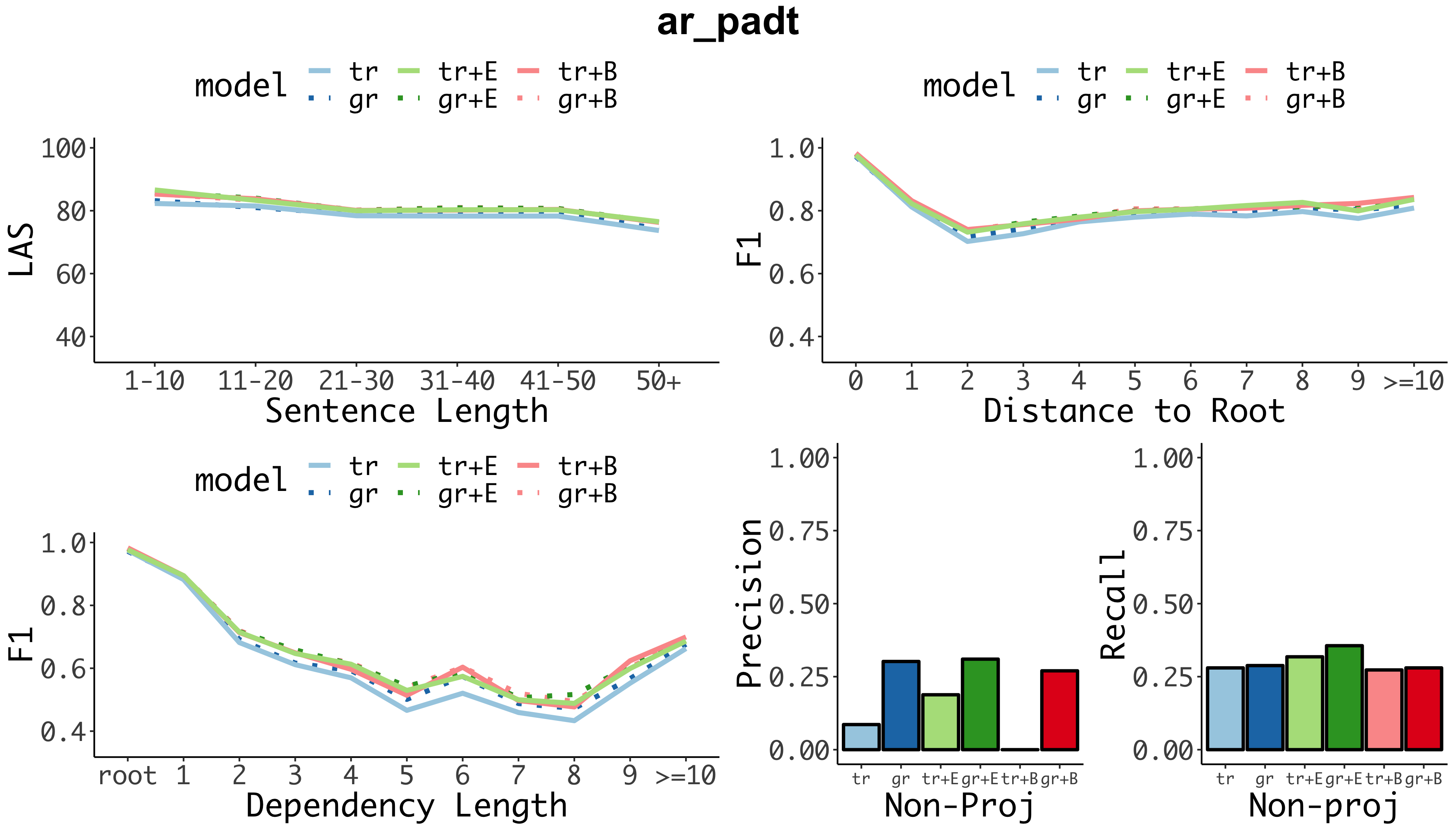}
\end{figure}

\begin{figure}[ht!]
\centering
\includegraphics[width=\columnwidth]{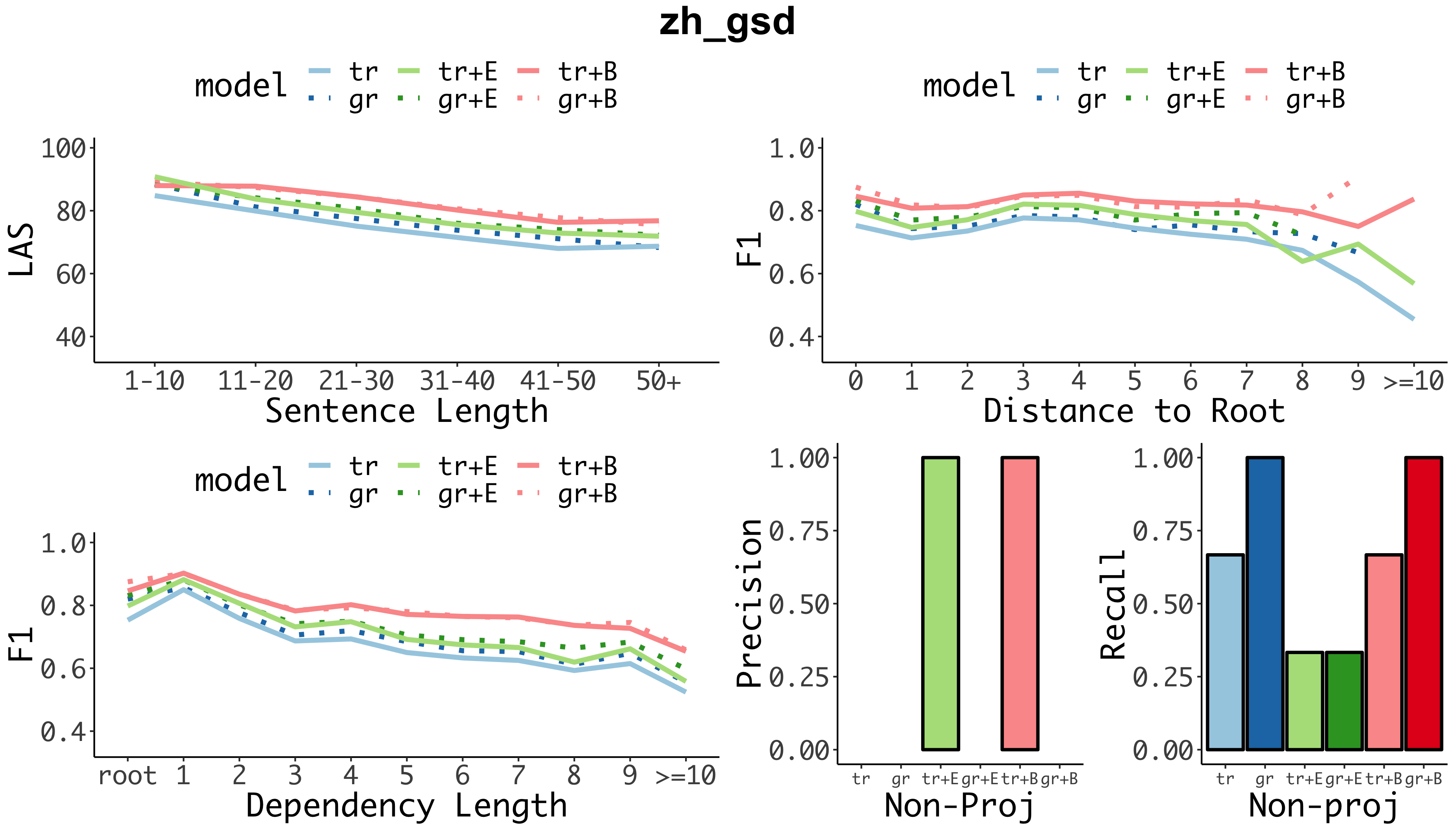}
\end{figure}

\begin{figure}[ht!]
\centering
\includegraphics[width=\columnwidth]{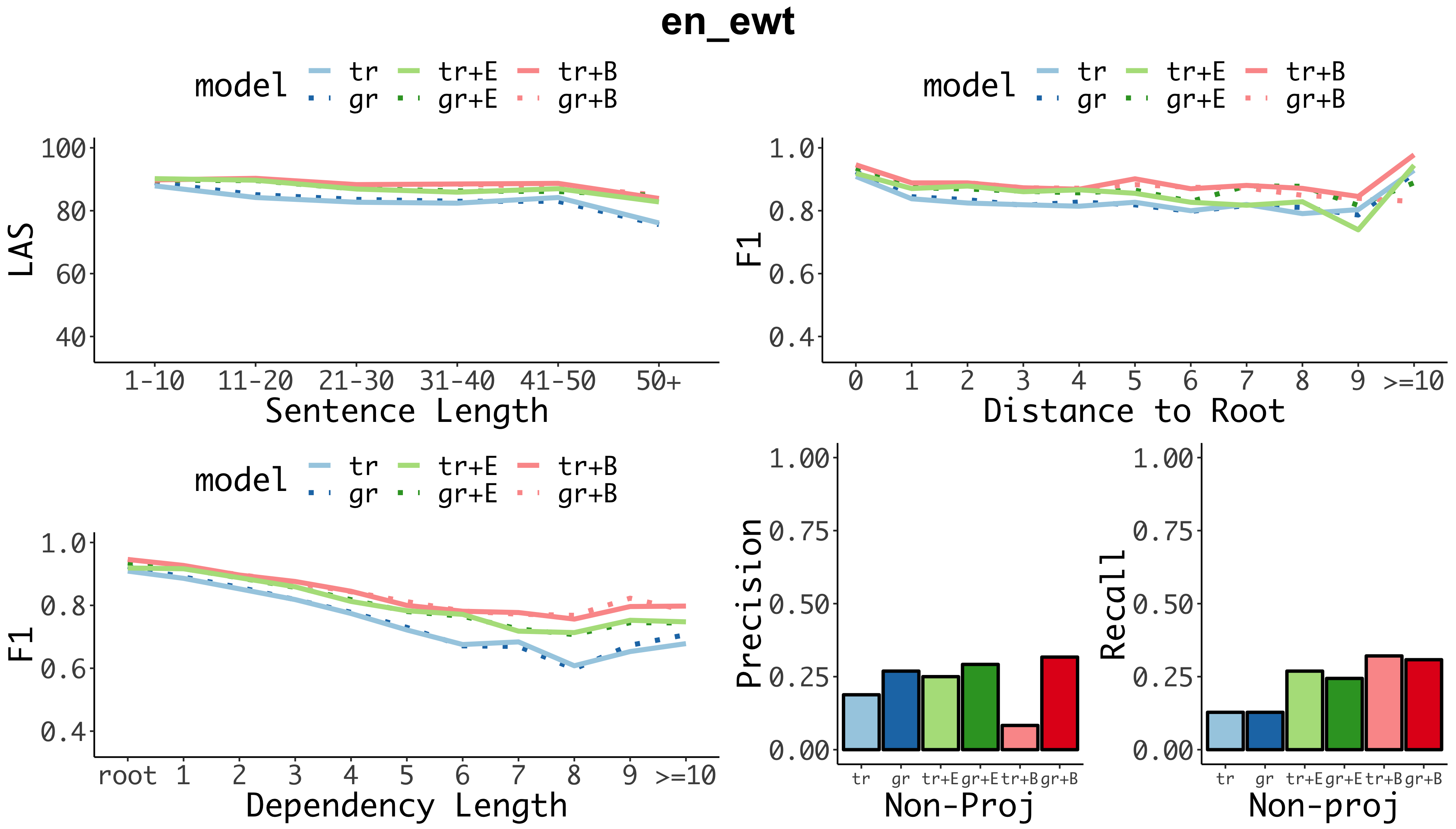}
\end{figure}

\begin{figure}[ht!]
\centering
\includegraphics[width=\columnwidth]{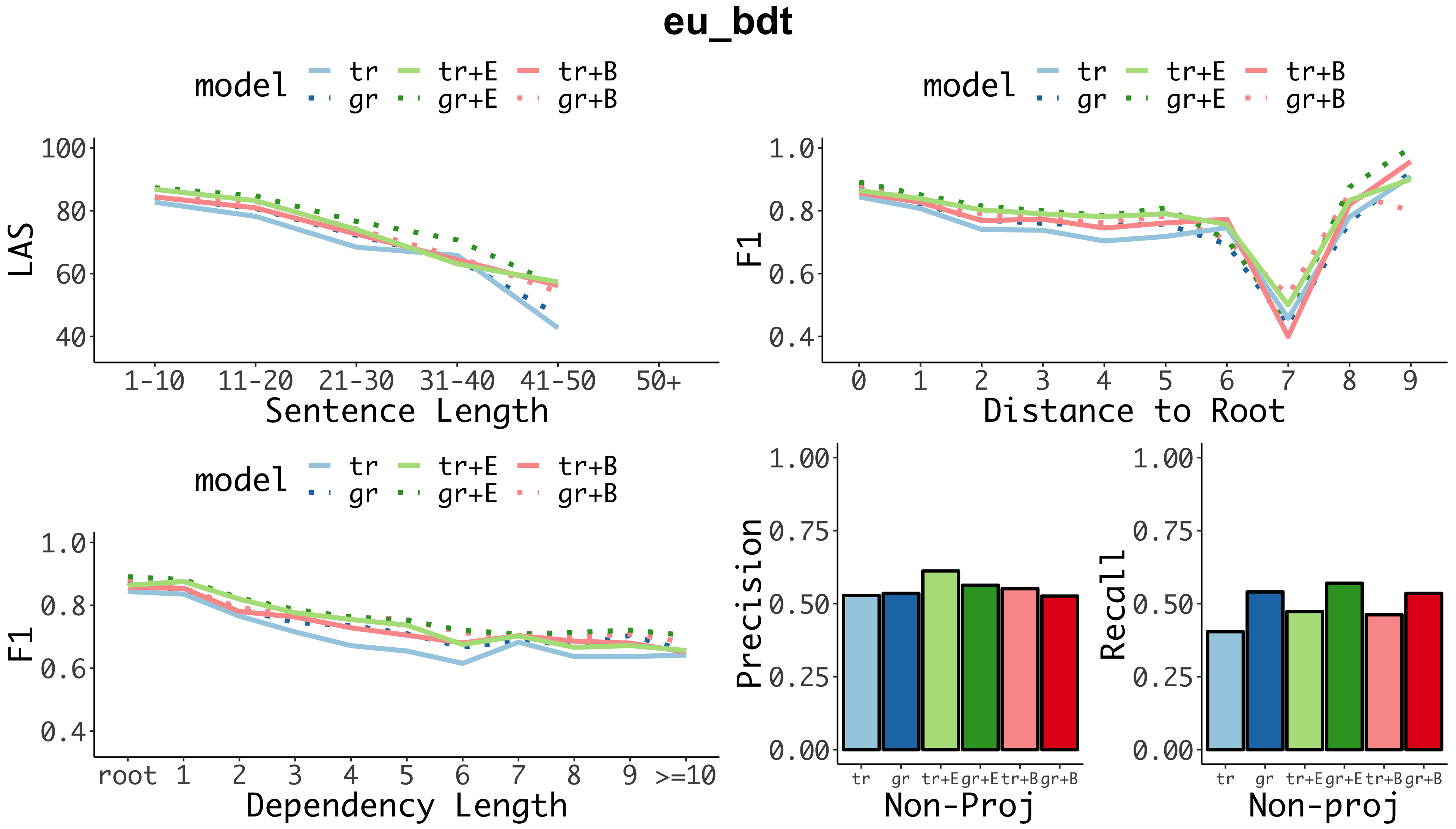}
\end{figure}

\begin{figure}[ht!]
  \onecolumn
\centering
\includegraphics[width=\columnwidth]{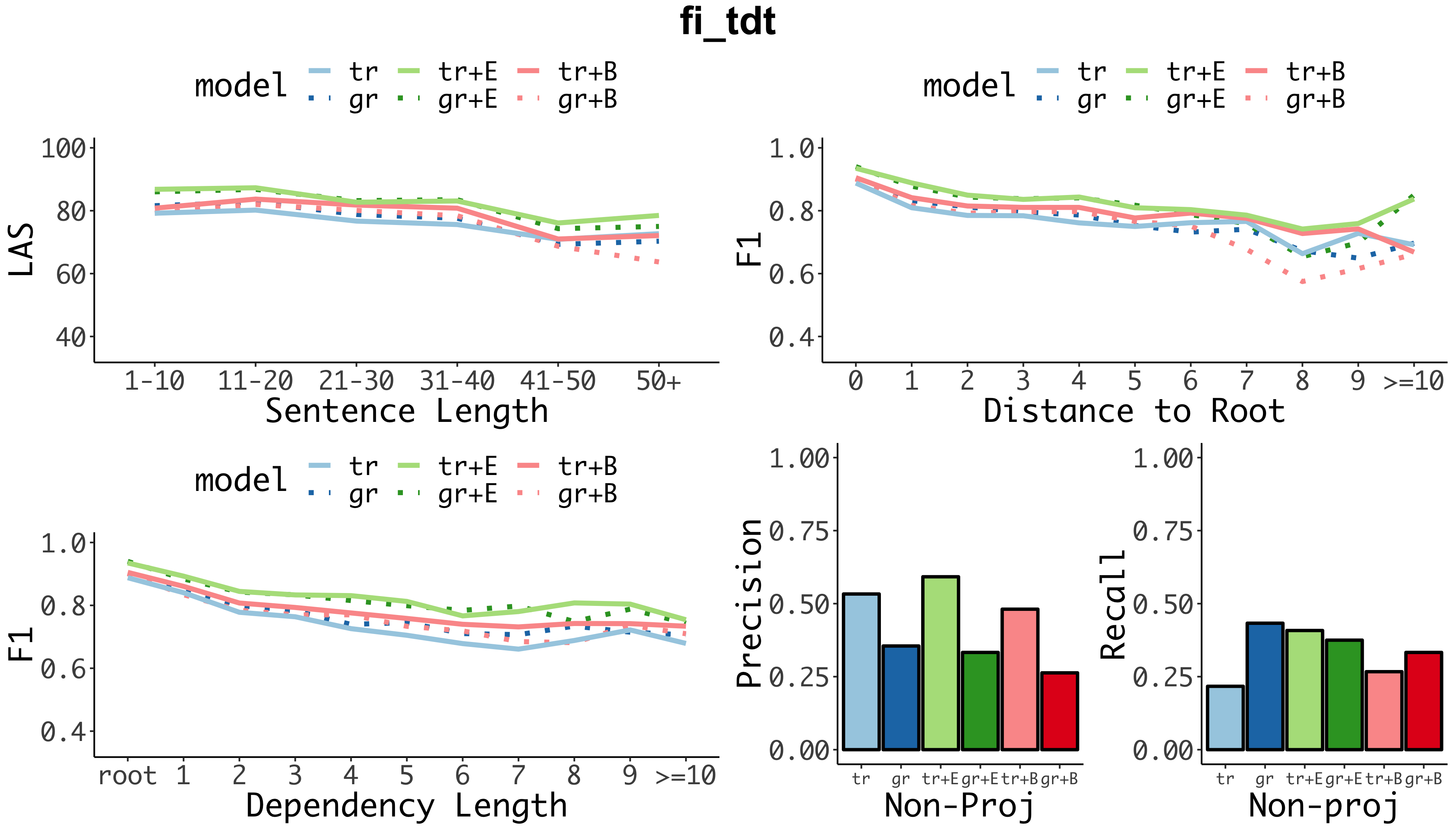}
\end{figure}

\begin{figure}[ht!]
\centering
\includegraphics[width=\columnwidth]{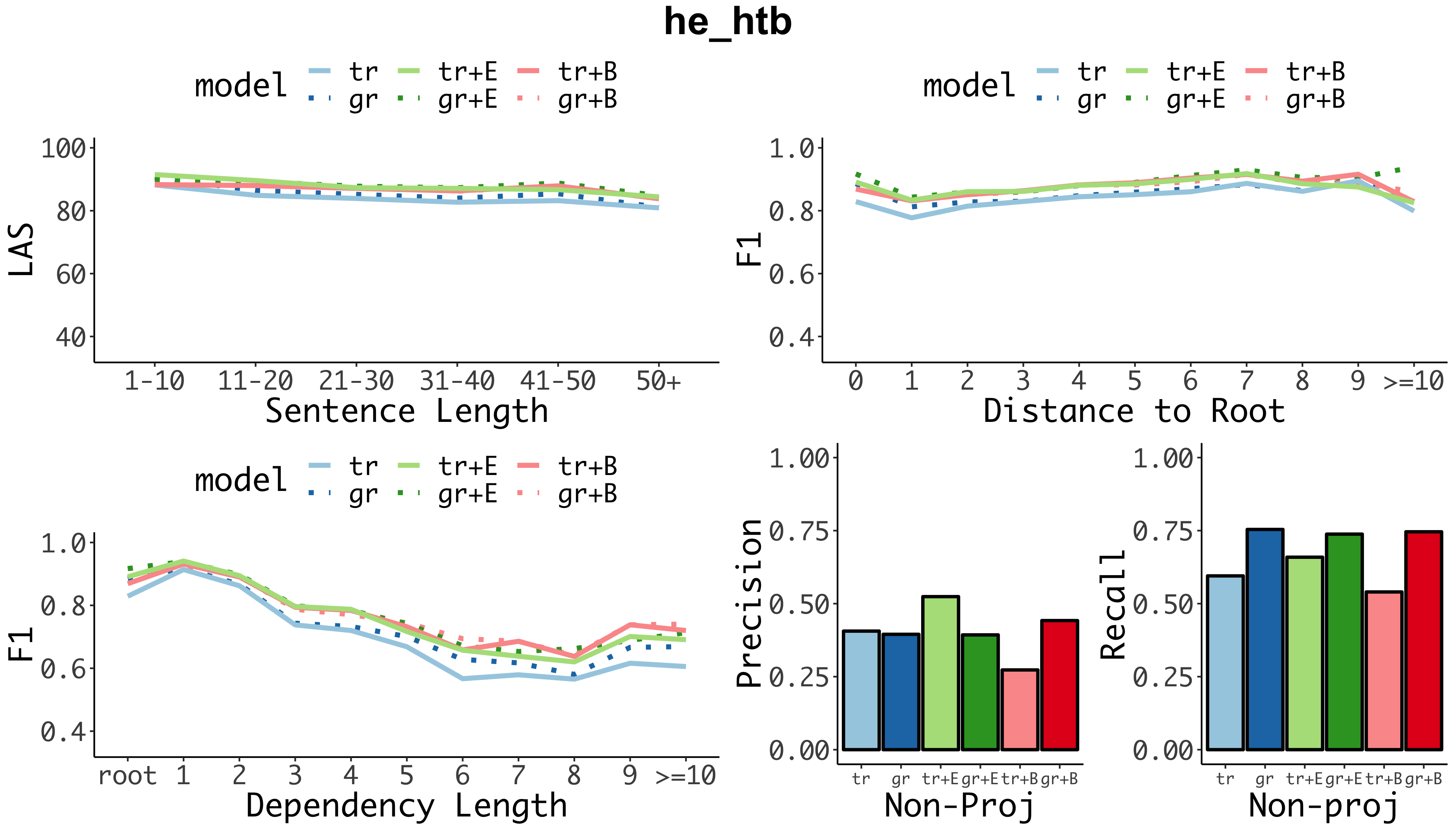}
\end{figure}

\begin{figure}[ht!]
\centering
\includegraphics[width=\columnwidth]{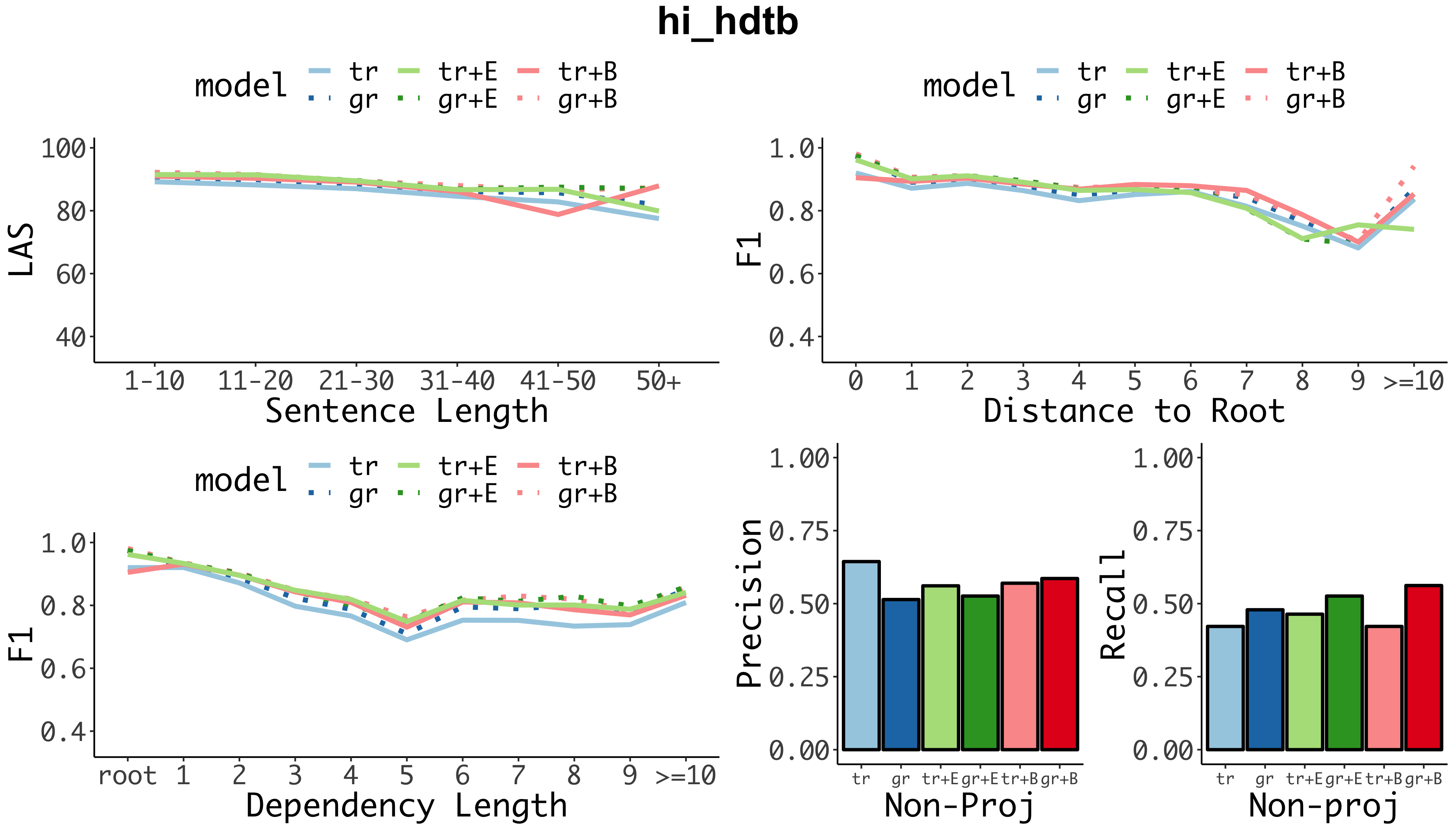}
\end{figure}

\begin{figure}[ht!]
\centering
\includegraphics[width=\columnwidth]{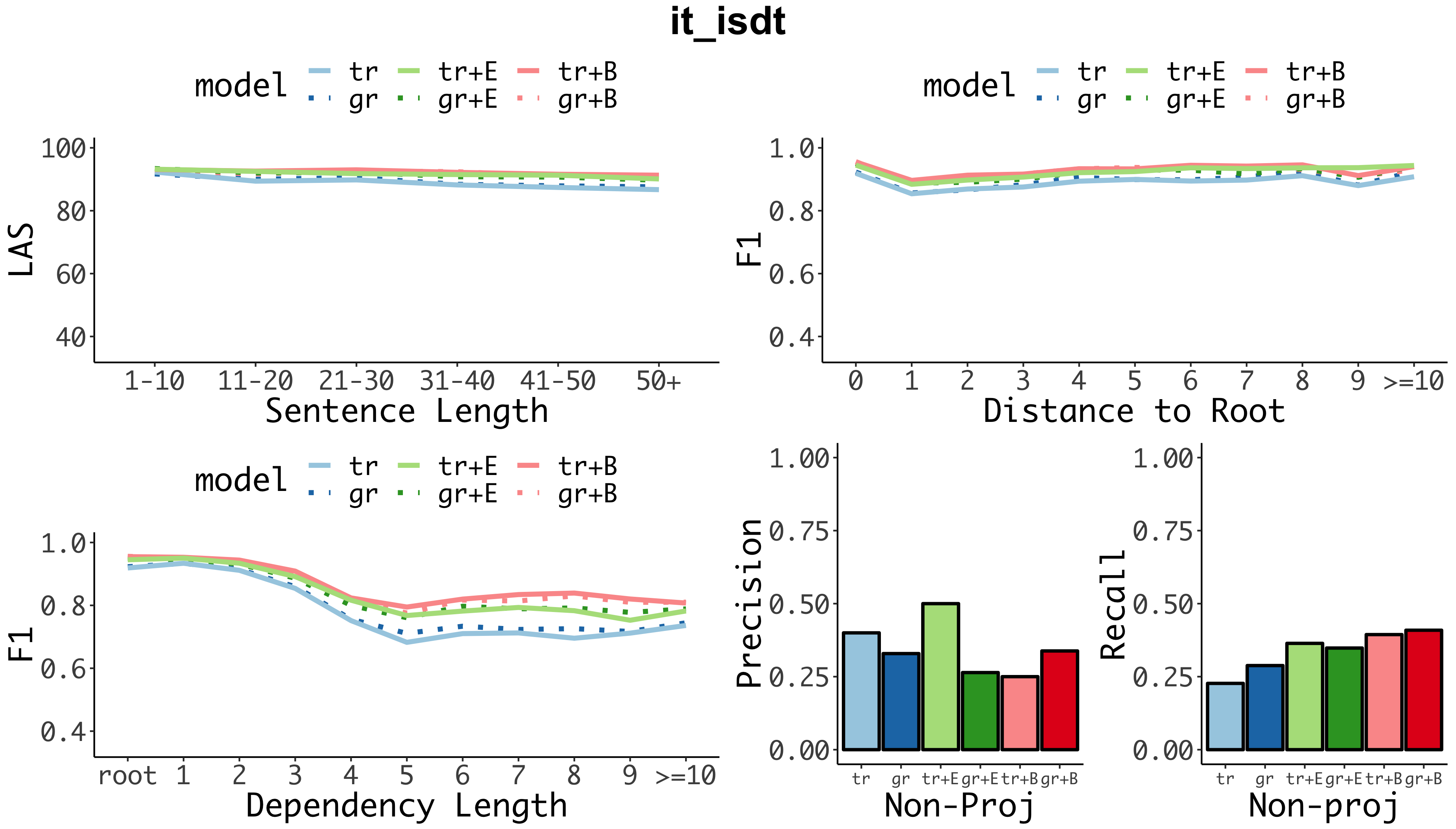}
\end{figure}

\begin{figure}[ht!]
\centering
\includegraphics[width=\columnwidth]{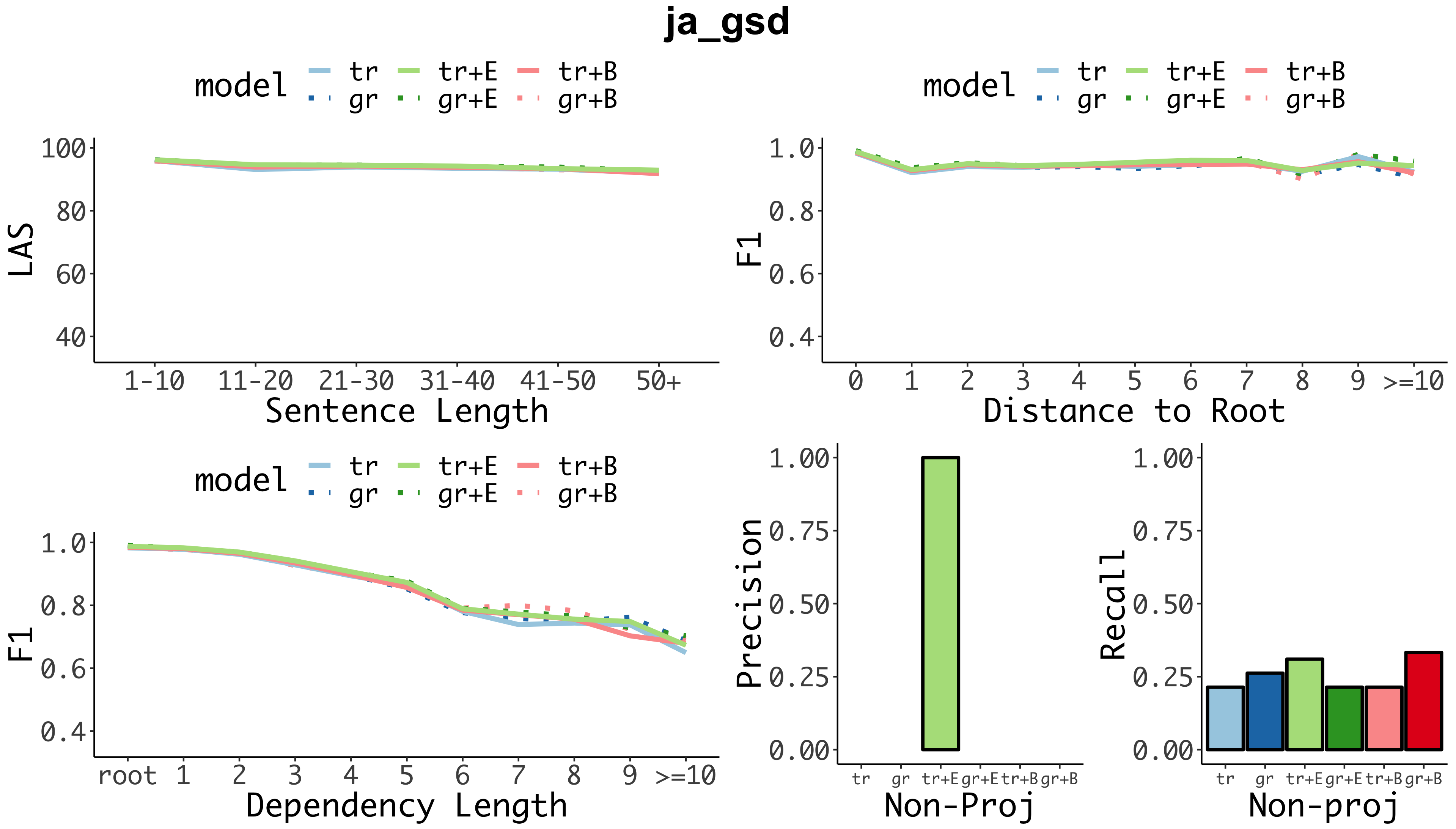}
\end{figure}

\begin{figure}[ht!]
\centering
\includegraphics[width=\columnwidth]{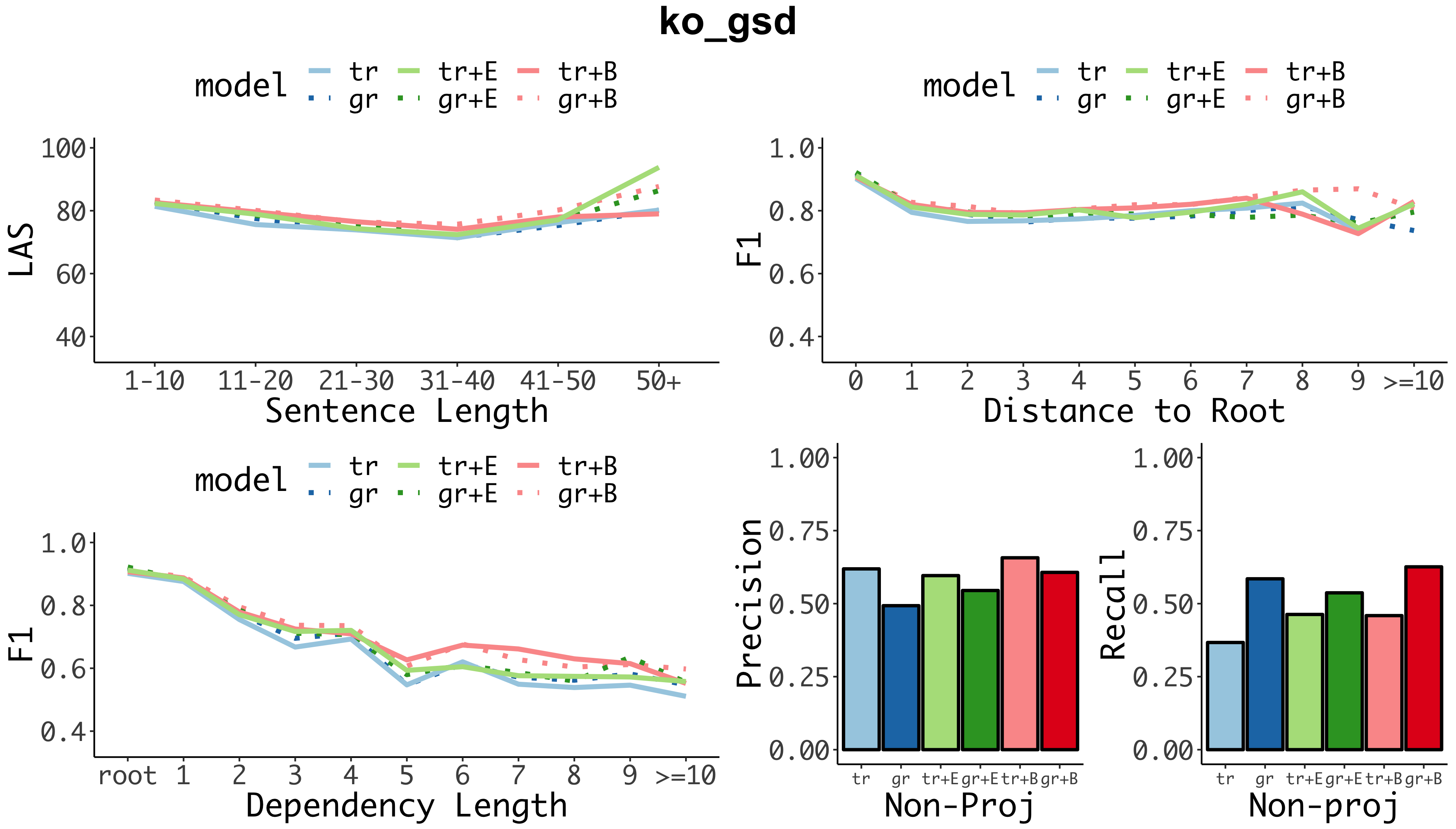}
\end{figure}

\begin{figure}[ht!]
\centering
\includegraphics[width=\columnwidth]{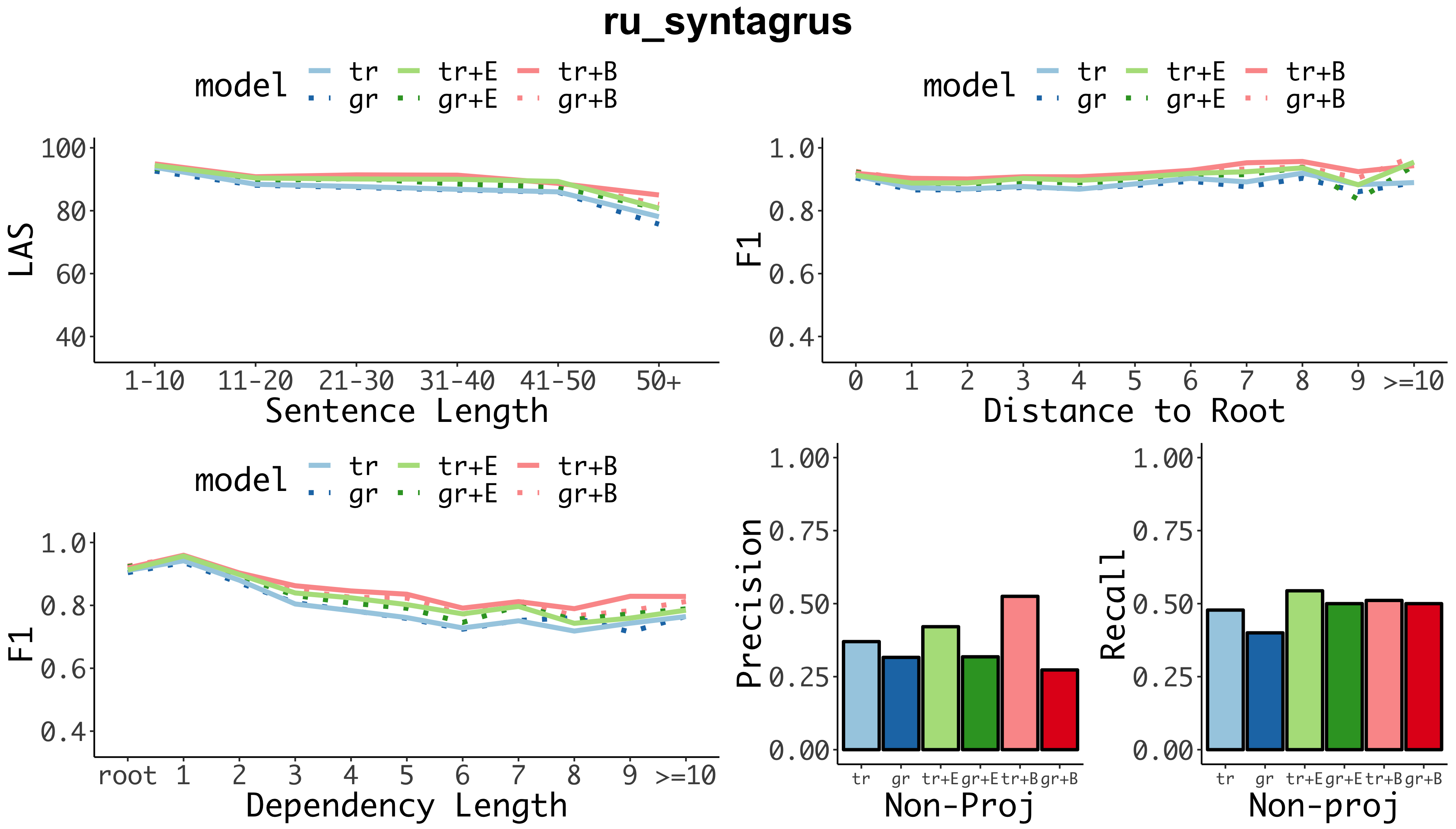}
\end{figure}

\begin{figure}[ht!]
\centering
\includegraphics[width=\columnwidth]{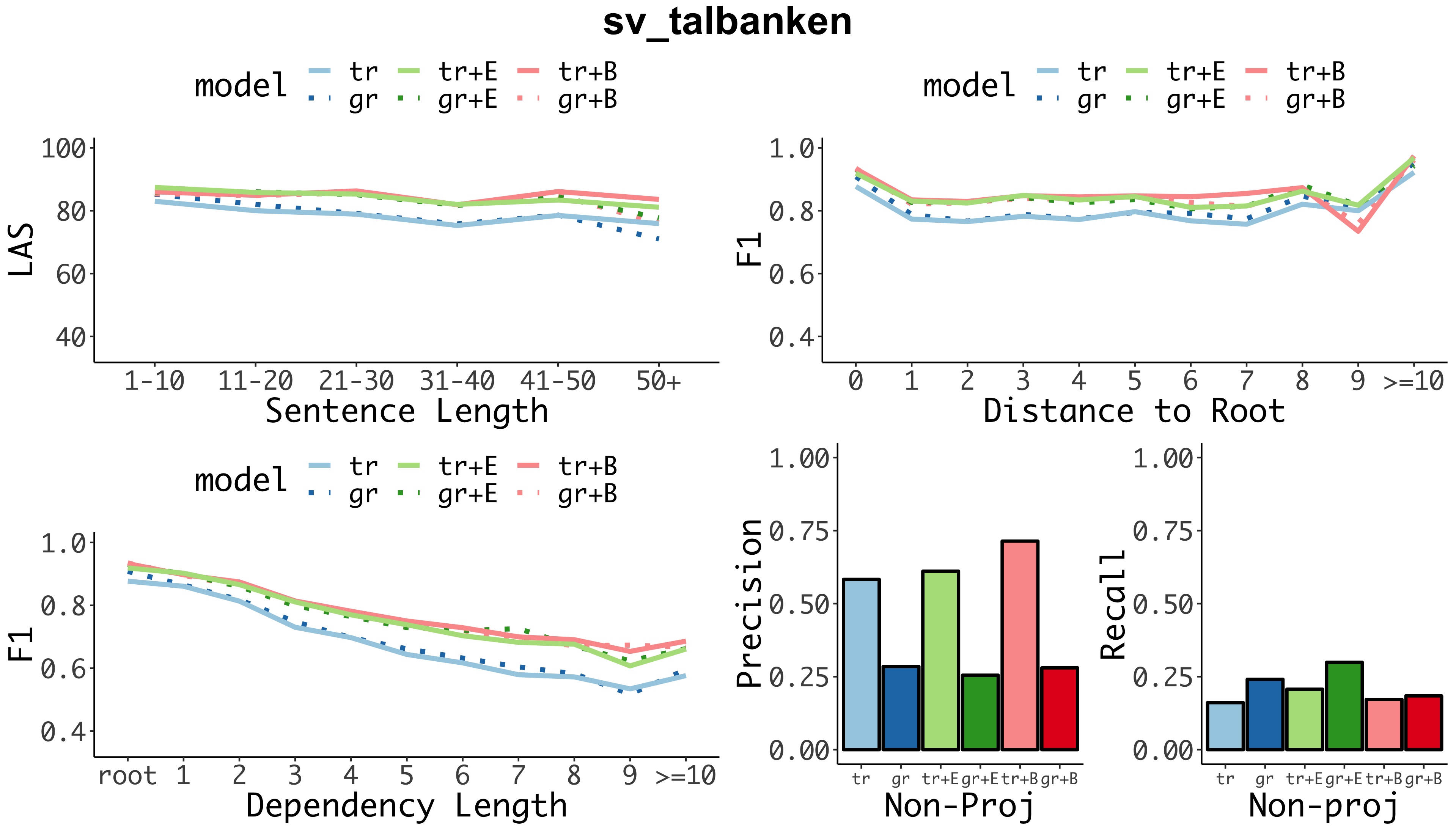}
\end{figure}

\begin{figure}[ht!]
\centering
\includegraphics[width=\columnwidth]{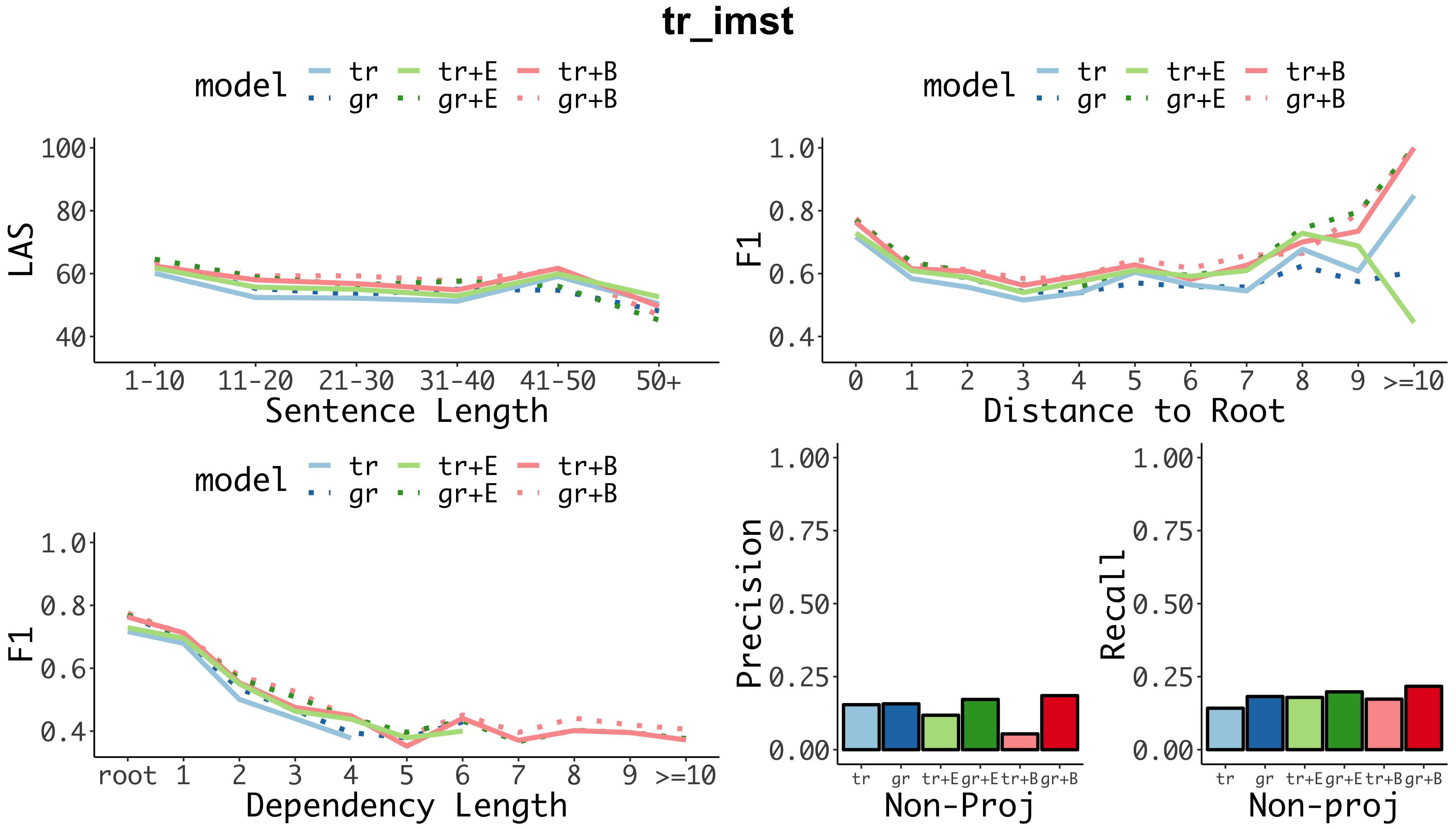}
\end{figure}


\end{document}